\documentclass[letterpaper]{article} % DO NOT CHANGE THIS
\usepackage[preprint]{arxiv2026_bridgealign}  % preprint mode: shows authors, suppresses copyright slug
\usepackage[hyphens]{url}  % DO NOT CHANGE THIS
\usepackage{graphicx} % DO NOT CHANGE THIS
\usepackage{natbib}  % DO NOT CHANGE THIS AND DO NOT ADD ANY OPTIONS TO IT
\usepackage{caption} % DO NOT CHANGE THIS AND DO NOT ADD ANY OPTIONS TO IT
\usepackage{algorithm}
\usepackage{algorithmic}

\usepackage{newfloat}
\usepackage{listings}
\DeclareCaptionStyle{ruled}{labelfont=normalfont,labelsep=colon,strut=off} % DO NOT CHANGE THIS
\floatstyle{ruled}
\newfloat{listing}{tb}{lst}{}
\floatname{listing}{Listing}

\usepackage{booktabs}

\usepackage{placeins}
\usepackage{tabularx} %表格自动换行
\usepackage{array}
\usepackage{multirow}
\usepackage{booktabs}
\usepackage{graphicx}
\usepackage{amsthm,amsmath,amssymb} %希腊字母
\usepackage{mathrsfs} %花体字
\usepackage{bm} %公式中的加粗
\usepackage{pifont} %公式需要
\usepackage{soul}
\usepackage{caption}
\usepackage{subcaption}
\usepackage{enumitem}
\usepackage{pdfpages}
\usepackage{dsfont}
\usepackage{color}
\usepackage{colortbl}
\usepackage{xcolor}
\usepackage{tcolorbox}
\tcbuselibrary{skins,breakable}
\usepackage{tikz}
\definecolor{deepred}{RGB}{180,0,0} % 深红色的RGB值
\definecolor{customred}{RGB}{170, 20, 20} % 替换为图片中的红色
\definecolor{customgreen}{RGB}{20, 170, 20} % 替换为图片中的绿色
\definecolor{warmgrayframe}{RGB}{118,126,136}
\definecolor{warmgrayback}{RGB}{241,244,247}
\definecolor{warmgraytitle}{RGB}{110,118,128}
\usepackage{xspace}
\newtcolorbox{promptbox}[1]{
  enhanced,
  rounded corners,
  arc=2mm,
  colframe=warmgrayframe,
  colback=warmgrayback,
  boxrule=0.8pt,
  top=4mm,
  bottom=4mm,
  left=4mm,
  right=4mm,
  width=\textwidth,
  fontupper=\ttfamily\small,
  title={#1},
  coltitle=white,
  colbacktitle=warmgraytitle,
  boxed title style={
    size=small,
    colframe=warmgrayframe,
    center title,
    before upper=\large\bfseries,
    boxrule=0.8pt,
  },
  overlay={
    \draw[warmgrayframe,line width=0.8pt]
      (frame.north west) -- (frame.north east);
  }
}
\usepackage{xurl} %会自动在所有可能的地方断行
\title{BridgeAlign: Bridging Preference Alignment for Humanities and Social Sciences}
\author{
    Ru Peng\textsuperscript{\rm 1}\thanks{This work was done during internships at Qwen Team, Alibaba Group and Inclusion AI, Ant Group.},
    Haokai Xu\textsuperscript{\rm 1},
    Xijun Gu\textsuperscript{\rm 1},
    Tianyu Zhao\textsuperscript{\rm 3},
    Zhiting Fan\textsuperscript{\rm 1},\\[3pt]
    Yawen Zeng\textsuperscript{\rm 1},
    Yihong Zhuang\textsuperscript{\rm 3},
    Jinyang Zhang\textsuperscript{\rm 2,\rm 4},
    Kexin Yang\textsuperscript{\rm 2},
    Jian Wu\textsuperscript{\rm 5},\\[3pt]
    Hao Chen\textsuperscript{\rm 1,\rm 2},
    Junyang Lin\textsuperscript{\rm 2},
    Dayiheng Liu\textsuperscript{\rm 2}\corresponding,
    Junbo Zhao\textsuperscript{\rm 1}\corresponding
}
\affiliations{
    \textsuperscript{\rm 1}Zhejiang University \quad
    \textsuperscript{\rm 2}Qwen Team, Alibaba Group \quad
    \textsuperscript{\rm 3}Inclusion AI, Ant Group\\[3pt]
    \textsuperscript{\rm 4}Peking University \quad
    \textsuperscript{\rm 5}Westlake University\\[3pt]
    \{rupeng, j.zhao\}@zju.edu.cn \quad
    liudayiheng.ldyh@alibaba-inc.com
}

\begin{document}

\maketitle

\begin{abstract}
While data synthesis for large language models (LLMs) is prevalent, it primarily targets domains with verifiable answers, overlooking open-ended humanities and social sciences~(HSS), where nuanced quality judgments matter more than objective correctness.
This makes preference alignment a natural paradigm for broad HSS tasks. Yet existing methods are either costly or not tailored to broad HSS disciplines. We thus propose \textbf{BridgeAlign}, among the first preference-alignment pipelines for broad HSS disciplines, with three phases:
i) \emph{Seed Curation}: curating HSS seed documents from web corpora via heuristic/LLM-based filtering and text refinement; 
ii) \emph{Preference Data Synthesis}: generating preference triplets via persona-based instruction inversion with Q\&A consistency checks; 
iii) \emph{Preference Optimization}: moving beyond naive human-vs-model heuristics by first grounding preferences in HSS quality rubric, then generating transitional responses via controlled quality degradation to form near-boundary preference pairs for finer-grained quality discrimination.
Aligning over 210k synthetic preference samples, BridgeAlign enables Qwen3-8B to achieve the best average across 17 benchmarks against 11 strong baselines---importantly, leading on both human-preference and knowledge-based capabilities at once, with no trade-off between them, as supported by extensive experiments and contextualized by existing theories\footnote{We will release all data, code, and models to foster community research.}.
\end{abstract}

\section{Introduction}
Data quality and diversity are crucial for large language models (LLMs). However, the prohibitive cost of human annotation~\cite{liu2024best} has rendered data synthesis a compelling alternative. Existing efforts primarily focus on verifiable tasks with standard answers, such as mathematics~\cite{yu2024metamath}, code~\cite{austin2021program}, tool-use~\cite{cai2023large}, and tabular data~\cite{zhao2025tabula}. In contrast, open-ended humanities and social sciences (HSS) tasks have long been overlooked. HSS data lacks standard answers, requiring nuanced human judgment for quality assessment rather than objective correctness. Consequently, the preference alignment paradigm~\cite{ouyang2022training,rafailov2024direct} is naturally suited to HSS tasks, motivating the need for preference data. However, existing preference datasets rely on human annotation~\cite{fan2018hierarchical} or costly proprietary models~\cite{cui2023ultrafeedback}, leaving data scarcity unresolved.
Although methods such as Self-rewarding~\cite{yuan2024self} use LLMs to self-evaluate their responses for iterative direct preference optimization (DPO) training, and Magpie~\cite{xu2024magpie} induces instruction generation via chat templates to build preference pairs from scored responses, they overlook HSS scenarios and fail to model the fine-grained, human-like judgments they require.
Meanwhile, HSS-related works like Weaver~\cite{wang2024weaver} and PersonaHub~\cite{ge2024scaling} focus on narrower creative writing or role-playing, rather than broader HSS domains. \textbf{To our knowledge, little work has explored synthetic preference alignment across broad HSS disciplines.}

\begin{figure*}[!t]
\centering
\includegraphics[width=1.\textwidth]{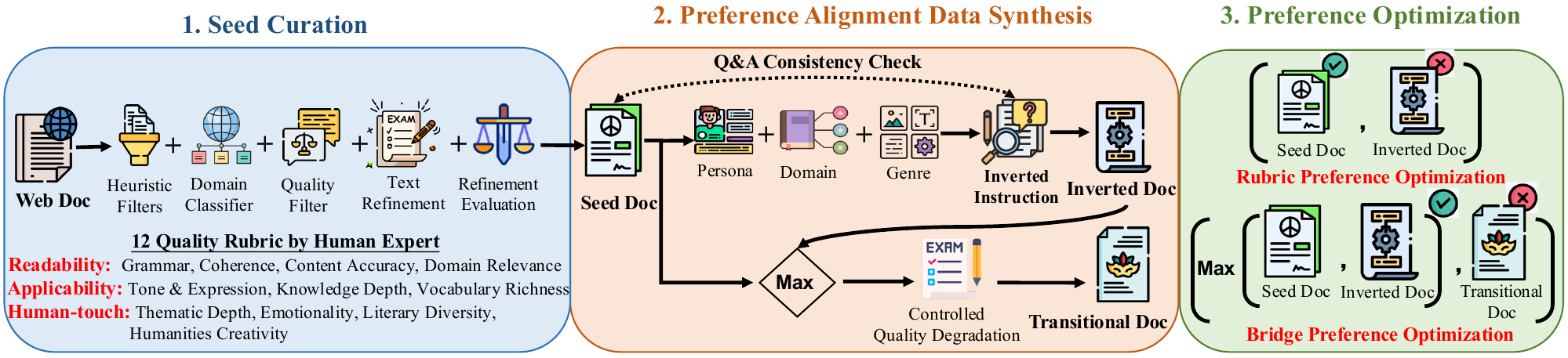}

\caption{
Overview of BridgeAlign: (1) \emph{Seed Curation}: filter and refine web texts to curate HSS seed documents; (2) \emph{Preference Alignment Data Synthesis}: generate inverted instruction and document from the seeds, use a Q\&A consistency check to filter out low-fidelity instructions, and controllably degrade the higher-scoring document (from either seed or inverted) to produce transitional documents; (3) \emph{Preference Optimization}: build pairwise preference data, then optimize model alignment via RubricPO and BridgePO.
}

\label{fig:method}
\end{figure*}

To address this gap, we follow official subject taxonomies\footnote{We identify all HSS disciplines from the \emph{Arts \& Humanities and Social Sciences} categories in the widely recognized QS subject taxonomies: \url{https://www.topuniversities.com/subject-rankings}} to focus on 14 HSS domains and propose BridgeAlign (see Figure~\ref{fig:method}), among the first preference alignment pipeline for broad HSS disciplines.
BridgeAlign aims to construct a sufficient-scale dataset (over 210K samples) with broad domain and content diversity, high multidimensional quality, and nuanced preference distinctions, so it requires multiple stages: 
1) We curate clean \emph{seed documents} from web resources via heuristic filtering, HSS classification, expert-crafted quality rubrics, and text refinement equipped with evaluation. 
2) We then apply persona-based instruction inversion to generate diverse instructions and corresponding inverted documents, followed by Q\&A consistency checks to ensure instruction fidelity, forming preference triplets $\langle\text{instruction, seed document, inverted document}\rangle$.
However, naively treating human-written as preferred and model-generated as rejected on these triplets—namely \textbf{H-MPO}—easily causes models to exploit superficial source cues rather than true quality. 
3) To overcome this empirical heuristic, we introduce rubric-based preference optimization (\textbf{RubricPO}), which grounds preferences in HSS quality rubric scores. 
However, RubricPO reveals a notable quality gap: seed documents often contain web noise, while inverted documents can be overrated due to model affinity effects~\cite{li2025preference}. 
Such easily separable pairs (``easy'' negatives) provide a limited signal for learning fine-grained quality features.
We therefore introduce the bridge preference optimization (\textbf{BridgePO}), the core of BridgeAlign, which controllably degrades the higher-scoring one between seed and inverted documents along the human-touch dimension to create \emph{transitional documents}, yielding harder-to-distinguish, near-boundary ``hard'' negatives that encourage the model to capture nuanced quality differences.

Experimentally, we train BridgeAlign on Qwen3-8B, Llama3.1-8B-Instruct, and Qwen2.5-14B-Instruct, and evaluate on 17 mainstream benchmarks covering nine core LLM capabilities. On Qwen3-8B, BridgeAlign variants achieve the best average performance against 11 strong baselines, leading on both human-preference and knowledge-based tasks at the same time.
{To verify robustness, we conduct comprehensive ablations on preference optimization variants, model architectures and scales, and data scaling effects. At the data level, BridgeAlign synthetic data outperforms the baselines in text length, lexical richness, and semantic diversity; on HSS tasks, the preference alignment paradigm also consistently outperforms the instruction-tuning paradigm. The Appendix further reports controlled tests on token budget and HSS specificity, and draws on existing theories of LLM data synthesis and DPO to help explain the observed experimental patterns.}
{Finally, using our HSS quality rubrics, we observe that the aligned models capture fine-grained quality features, achieving consistent but modest gains in the human-touch dimension.}
Overall, our contributions are as follows:
\begin{itemize}%[leftmargin=*,itemsep=0pt,topsep=2pt]
\item To our knowledge, we present \textbf{among the first preference data synthesis and optimization work for broad HSS disciplines}, filling a critical void in LLM alignment.
\item We propose the \textbf{novel BridgePO paradigm, which applies rubric-guided controlled quality degradation to high-quality samples to construct near-boundary \emph{hard} negatives} \textbf{for fine-grained HSS alignment}, with RubricPO as an intermediate product.
\item BridgeAlign \textbf{achieves the best average across 17 benchmarks covering 9 LLM capabilities}, with its efficacy validated by extensive empirical analysis and interpreted by existing theories.
\end{itemize}

\section{Related Work}

\paragraph{Open-domain tasks for LLM} unlike verifiable ones, lack standard answers and require nuanced human judgment, making data collection and synthesis challenging.
Existing work is limited and focuses on narrow aspects: 1)-\emph{Writing}: simple story generation~\cite{eldan2023tinystories}, creative writing~\cite{wang2024weaver}; 2)-\emph{Dialogue systems}: task-oriented dialogue for e-commerce~\cite{qian2025bottom}, few-shot dialogue summarization~\cite{lu2025mutual}, multi-turn multi-topic dialogues~\cite{lee2025doctalk}; 3)-\emph{Role playing}: 1 billion persona construction~\cite{ge2024scaling}; and 4)-\emph{Long context}: ultra long-context generation~\cite{bai2024longwriter}. Overall, open-domain research remains underexplored and fragmented.

\paragraph{Preference Alignment} guides LLMs toward human-preferred outputs, and mainly involves preference-data generation and preference optimization.
For data generation, representative works include human-crafted datasets such as \emph{WritingPrompts}~\cite{fan2018hierarchical}, which pairs Reddit prompts with human-written stories; 
mixed datasets such as \emph{Ultrafeedback}~\cite{cui2023ultrafeedback}, which aggregate open-source prompts and multi-model responses evaluated by GPT-4; 
and synthetic approaches such as Constitutional AI~\cite{bai2022constitutional} and RLAIF~\cite{lee2023rlaif}, which generate preference signals without direct human labels. Recent methods such as \emph{SynthQuestions}~\cite{zhu2025real} and \emph{Magpie}~\cite{xu2024magpie}, synthesize data via few-shot prompting or chat-template prompting, while \emph{Self-Rewarding}~\cite{yuan2024self} and \emph{FollowSoftConstraint}~\cite{ren2025step} improve alignment data quality via self-iteration or progressive soft constraints.
For preference optimization, DPO optimizes pairwise preference loss, LiPO~\cite{liu2024lipo} uses listwise rankings for finer alignment, and Zephyr~\cite{tunstall2023zephyr} applies distilled DPO with AI feedback. Selective DPO~\cite{gao2025principled} filters overly difficult samples based on validation loss, Beta-DPO~\cite{wu2024beta} tunes the $\beta$ parameter to accommodate different preference gaps, and IPO~\cite{azar2024general} penalizes log probabilities to avoid premature convergence.
Building on these efforts, BridgeAlign integrates data synthesis with tailored preference optimization algorithms to improve LLM performance on HSS tasks.

\section{BridgeAlign}

We introduce BridgeAlign in Figure~\ref{fig:method}, a three-phase pipeline for synthesizing preference alignment data for the humanities and social sciences. The framework comprises: \emph{seed curation}, \emph{preference data synthesis}, and \emph{preference optimization}, as described below.

\subsection{Seed Curation}\label{sec:seed_curation}

Following prior works that leverage vast web data \cite{yue2024mammoth2,li2024selfalignment}, we select the 627B-token SlimPajama corpus \cite{cerebras2023slimpajama} to curate seed data.
We first tokenize and segment the documents into 200 to 4k tokens to match real-world length distributions. 
To mitigate prevalent web noise and the scarcity of high-quality HSS content, we filter out non-HSS sources (e.g., StackExchange, GitHub), retain all documents from Books, ArXiv, and Wikipedia, and sample 10\% from C4 and CommonCrawl, yielding an initial pool of 30M documents. We then apply the \texttt{datatrove} toolkit\footnote{\url{https://github.com/huggingface/datatrove}} with multiple heuristic filters such as fastText~\cite{joulin2016fasttext}, Gopher~\cite{rae2021scaling}, C4~\cite{raffel2020exploring}, Fineweb~\cite{penedo2024fineweb}, and MinHash~\cite{broder1997resemblance} to remove noise and redundancy. 
Next, we employ an LLM to categorize the filtered texts into 14 HSS domains defined by the official subject taxonomy.
{We further introduce 12 expert-crafted quality rubrics to prompt the LLM to filter documents meeting specific thresholds: readability ($=5$), applicability ($\geq4$), and human-touch ($\geq3$) (details in Appendix~\ref{sec:hss_quality_rubrics}).}
Even after several processes, the resulting documents still contain noise or poor expressions. We thus employ an LLM-based refinement to cleanse content and optimize expression while preserving fidelity. A final rigorous refinement evaluation ensures all refined texts satisfy the above three requirements to serve as clean seed documents.

\subsection{Preference Alignment Data Synthesis}\label{sec:data_synthesis}

\paragraph{Instruction Inversion}\label{sec:instruct_inversion}
Inspired by previous methods~\cite{koksal2023longform,li2024selfalignment}, we reverse-generate instructions from seed documents by setting explicit and actionable constraints, guiding LLMs to faithfully reproduce the original seed documents.
Each inverted instruction specifies the domain, genre, and length of the seed document, summarizes core content, and outlines text structure and narrative perspective, while remaining concise, non-redundant, and avoiding direct references such as \textit{``source document''}.
To enhance diversity, we incorporate persona settings (stance, mindset, tone) and uniformly adopt second-person expressions. 
Unlike previous approaches that require predefined persona hubs, knowledge bases, or numerous task templates~\cite{ge2024scaling,li2024synthetic,nayak2024learning}, our method is more flexible and robust.
This instruction inversion yields diverse inverted instructions, taking prompt diversity into account when constructing preference data~\cite{li20252d}.

\paragraph{Q\&A Consistency Check}\label{sec:qa_consistency_check}
Next, to ensure the fidelity of instructions reverse-generated from seed documents, we introduce a Q\&A Consistency Check. The LLM first answers each inverted instruction to output the corresponding inverted document, and then performs a rationale-backed binary judgment based on its ``textual consistency'' (alignment of core content and key points) and ``instruction effectiveness'' (coverage of core information without explicit citation) between the seed and inverted documents, assessing whether the instruction faithfully reproduces the seed document.
Only when both criteria are met can the high fidelity of the inverted instruction be ensured, retaining the triplet of $\langle$inverted instruction, seed document, and inverted document$\rangle$.

\subsection{Preference Optimization}\label{sec:preference_optimization}

\paragraph{Rubric-based Preference Optimization}\label{sec:rubricPO}
Given the triplet $\langle$inverted instruction $I$, seed document, inverted document$\rangle$, we start from the empirical prior that human-written texts usually outperform model-generated outputs. Accordingly, the seed document is labeled \textit{chosen} $O_{w}$ and the inverted document \textit{rejected} $O_{l}$, forming the human–model preference pair $(I, O_{w}, O_{l})\!\in\!D_{\text{HM}}$. 
However, the distinct distribution gap between these sources causes the loss in \textbf{human–model preference optimization (H-MPO)} to converge rapidly, leading the model to exploit superficial source cues rather than truth quality signals.
To curb this bias, we reuse an expert-crafted HSS quality rubric to score document pairs. By assigning the higher-scoring text as \textit{chosen} $O_{w}$ while the lower-scoring one as \textit{rejected} $O_{l}$, we construct rubric preference pairs $(I, O_{w}, O_{l})\!\in\!D_{\text{Rubric}}$.
RubricPO adopts the same DPO loss with inverse-temperature hyper-parameter $\beta$ and logistic sigmoid $\sigma(\cdot)$, which is defined as $\mathcal{L}_{\text{RubricPO}}(\pi_\theta; \pi_{\text{ref}})$:
\begin{equation}
\resizebox{\columnwidth}{!}{ % 缩放至单栏宽度，避免溢出页边
$
\begin{aligned}
-\mathbb{E}_{(I, O_w, O_l) \sim \mathcal{D}} \Bigl[ \log \sigma \Bigl( 
&\underbrace{\beta \log \frac{\pi_\theta(O_w | I)}{\pi_{\text{ref}}(O_w | I)}}_{\text{
Implicit reward (pref.)} \uparrow}
- \underbrace{\beta \log \frac{\pi_\theta(O_l | I)}{\pi_{\text{ref}}(O_l | I)}}_{\text{Implicit reward (rej.)} \downarrow} \Bigr) \Bigr],
\end{aligned}
$
}\label{eq:dpo_objective}
\end{equation}
where $\pi_{\theta}$ denotes the current policy model and $\pi_{\text{ref}}$ is the frozen reference model, and $\pi(O|I) \in [0, 1]$ is the conditional probability of generating completion $O$ given prompt $I$. By aligning directly with textual HSS quality, RubricPO better satisfies human preferences and knowledge understanding compared with H-MPO (see Sec.~\ref{RubricPO_vs_H-MPO} for details).

\paragraph{Validation of HSS Quality Rubric} 
Before use, we validated the HSS quality rubric on 400+ inverted documents that varied in length, source, domain, and quality levels. These documents were shuffled and assigned to 20 human annotators, who independently rated each rubric on a 1–5 scale. Following the LLM-as-a-judge framework \cite{zheng2023judging}, we then evaluated the documents using Qwen3-30B-A3B (our data synthesis model) and GPT-4.1 (a strong judge model), yielding 86\% and 91\% agreement with majority human ratings. Combined with high inter-annotator reliability measured by Kappa scores \cite{mchugh2012interrater}, these dual human and LLM assessments confirm the effectiveness of our HSS quality rubric.

\paragraph{Bridge Preference Optimization}\label{sec:BridgePO}
Although RubricPO performs well, a clear quality gap remains between seed and inverted documents, with seeds typically scoring lower (Figure \ref{fig:inv_seed_score_dist}). This gap arises from: (i) inherent web noise in seeds, and (ii) overrating of inverted documents due to ``model affinity,'' since they are generated by the same model as the evaluator \cite{li2025preference}. As a result, the chosen--rejected pairs are often easily separable, yielding ``easy'' negatives. This limitation is also common in prior degradation-based synthesis methods, which derive negatives from clearly separated good and bad responses, making it difficult to learn subtle quality differences \cite{stiennon2020learning}.

To construct preference samples near the decision boundary, we use rubric scores to select the higher-scoring one between the seed and inverted documents as the degradation source, and apply \emph{controlled degradation}: prompting the LLM to lower its four human-touch sub-scores (literary diversity, emotionality, thematic depth, and humanities creativity) toward lower targets while keeping readability and applicability unchanged. The resulting transitional documents serve as a ``bridge'' that narrows the quality gap. We then apply CodecLM-style filtering \cite{wang2024codeclm} to retain transitional documents whose scores fall within a predefined margin of the chosen ones. We train BridgePO on these chosen--transitional pairs to align all three quality dimensions and enable finer-grained, more human-like alignment. BridgePO's key is thus not degradation itself, but rubric-guided construction of near-boundary hard negatives.

\section{Experiment}

\subsection{Experimental Setup}

\paragraph{Baselines}
For robust evaluation, we compare the BridgeAlign synthetic dataset with 10 leading baselines, each containing 210k samples\footnote{Use all data if the dataset has under 210k samples.}. These encompass:
(1) Human-written datasets like \textbf{WritingPrompts}~\cite{fan2018hierarchical}; 
(2) Mixed-source datasets like \textbf{Ultrafeedback}~\cite{cui2023ultrafeedback}; 
(3) Synthetic datasets including \textbf{SynthQuestions}~\cite{zhu2025real} and \textbf{Magpie}~\cite{xu2024magpie} air, pro, mixed version. Furthermore, we reproduced representative methods under the same setup of seed data and synthesis model, including \textbf{Self-Rewarding}~\cite{yuan2024self} (+M$_2$ and +M$_3$ for the stages-1 and stages-2 DPO) and \textbf{FollowSoftConstraint}~\cite{ren2025step} (with and without curriculum learning).

\paragraph{Data Synthesis and Training Settings}
We synthesize data at the scale of 30M web documents (41B tokens) using the default generation settings of Qwen3-30B-A3B~\cite{yang2025qwen3}, balancing representation capacity and computational cost (single-GPU deployable). After filtering JSON parsing failures and removing \texttt{<thinking>} content, we obtain 210k preference samples and construct standard pairwise formats for alignment training. Following standard practices~\cite{qwen2.5,dubey2024llama}, we apply 13-gram exact matching to detect test-set leakage, ensuring that no test instances or their paraphrases are used. The synthesis process costs 2.4k GPU hours, with a per-document cost of 0.28 seconds. For preference optimization, we train Qwen3-8B (non-thinking mode), Qwen2.5-14B-instruct~\cite{qwen2.5}, and LLaMA3.1-8B-instruct~\cite{dubey2024llama} on multiple datasets using identical hyperparameters for fair comparison.

\paragraph{Evaluation Benchmarks}
We adopt 17 benchmarks spanning 9 task categories:
\textbf{emotion perception} (BuzzBench and EQ-Bench3~\cite{eqbench3_repo_2025}),
\textbf{role-playing} (RoleBench~\cite{wang2023rolellm}, assessing both instruction and role generalization), 
\textbf{writing skill} (WritingBench~\cite{wu2025writingbench}, CreativeWriting-v3 and Judgemark-v2~\cite{paech2023eq}), 
\textbf{social interactions} (Social-IQa~\cite{sap2019socialiqa}, IQuiz\_EQ~\cite{chen2024tombench}), 
\textbf{instruction following} (IFEval~\cite{zhou2023instruction}, Collie~\cite{yao2023collie}), 
\textbf{world knowledge} (MMLU~\cite{hendrycks2020measuring}, including HSS subsets),
\textbf{commonsense reasoning} (HellaSwag~\cite{zellers2019hellaswag}, StoryCloze~\cite{mostafazadeh2016corpus}), 
\textbf{long context} (CoQA~\cite{reddy2019coqa}, GovernmentReport\_CRS~\cite{shaham2022scrolls}), and 
\textbf{reading comprehension} (NarrativeQA~\cite{kovcisky2018narrativeqa}, Xsum~\cite{narayan2018don}). 
We further use AlpacaEval 2~\cite{dubois2024length} (GPT-4.1 judge) to measure win rate against Qwen3-8B, and build a blind pairwise human evaluation benchmark to validate the gains beyond LLM judges. To reduce variance, judge-based benchmarks are averaged over multiple runs (see Appendix~\ref{sec:benchmarks} for details).

\begin{table*}[t]
  \centering
  \renewcommand\arraystretch{1.1}
  \setlength{\tabcolsep}{1pt}
  \resizebox{1.0\textwidth}{!}
  {
  \begin{tabular}{llccccccccccccc}
    \toprule
    \multicolumn{1}{l}{\multirow{2}{*}{\textbf{Selected Method}}} &
    \multicolumn{1}{l}{\multirow{2}{*}{\textbf{Data Synthesis Model}}} &
    \multicolumn{5}{c}{\textbf{Human Preference}} &
    \multicolumn{6}{c}{\textbf{Knowledge based}} &
    \multicolumn{1}{c}{\multirow{2}{*}{\textbf{Avg}}} &
    \multicolumn{1}{c}{\multirow{2}{*}{\textbf{AlpacaEval 2}}} \\
    
    \cmidrule(lr){3-7}\cmidrule(lr){8-13}
    
    \multicolumn{2}{l}{} 
    & \textbf{Emotion} 
    & \textbf{Role} 
    & \textbf{Writing} 
    & \textbf{Social} 
    & \textbf{Avg HP}
    & \textbf{Instruct} 
    & \textbf{World} 
    & \textbf{Common} 
    & \textbf{Long} 
    & \textbf{Read} 
    & \textbf{Avg KB} \\
    \midrule
    
    Qwen3-8B & - 
    & 49.20 & 66.56 & 64.52 & 66.63 & 62.04 
    & 56.13 & 74.90 & 76.49 & 47.65 & 35.22 & 59.61 & 60.70 & 50.00\\
    \midrule

    Ultrafeedback & GPT-4 
    & 49.50 & 68.90 & 64.23 & 64.96 & 62.15 
    & 57.09 & \textbf{75.15} & 77.03 & 51.49 & 35.71 & 60.73 & 61.37 & 51.55\\

    WritingPrompts & Human 
    & 48.48 & 67.35 & 65.00 & 69.67 & 62.89 
    & 57.43 & 74.83 & 76.92 & 51.52 & 35.55 & 60.67 & 61.67 & 52.30\\

    SynthQuestions & Llama3-70B-Instruct 
    & 51.01 & 70.92 & 66.06 & 69.86 & 64.64 
    & 56.57 & 75.04 & 77.20 & 51.22 & 35.51 & 60.55 & 62.39 & 50.92\\

    Magpie\_Air & Llama3-8B-Instruct 
    & 49.43 & 69.49 & 65.67 & 66.44 & 63.08 
    & 56.52 & 75.07 & 77.01 & 51.65 & 35.67 & 60.63 & 61.73 & 50.85\\

    Magpie\_Pro & Llama3.1-70B-Instruct 
    & 50.10 & 69.57 & 65.63 & 67.90 & 63.56 
    & 56.76 & 75.13 & 77.13 & 51.24 & 35.84 & 60.66 & 61.97 & 51.12\\

    Magpie\_Mixed & Llama3.1-70B-Instruct
    & 49.56 & 70.76 & 65.41 & 66.39 & 63.29
    & 57.74 & 75.13 & 77.00 & \textbf{51.76} & 35.95 & 60.94 & 62.00 & 52.78\\

    Self-Rewarding+M$_2$ & Qwen3-30B-A3B 
    & 49.81 & 66.07 & 65.22 & 66.32 & 62.23 
    & 57.39 & 74.79 & 76.68 & 51.65 & \textbf{36.24} & 60.75 & 61.42 & 52.15\\

    Self-Rewarding+M$_3$ & Qwen3-30B-A3B 
    & 51.13 & 69.25 & 66.27 & 67.23 & 63.78 
    & 58.08 & 75.09 & 76.73 & 51.68 & 35.28 & 60.80 & 62.14 & 53.15\\

    FollowSoftConstraint w/ CL & Qwen3-30B-A3B 
    & 49.02 & 65.92 & 63.98 & 69.25 & 62.26 
    & 56.30 & 75.06 & 76.61 & 51.63 & 35.18 & 60.42 & 61.25 & 50.65\\

    FollowSoftConstraint w/o CL & Qwen3-30B-A3B 
    & 49.00 & 66.00 & 63.77 & 68.17 & 61.96 
    & 57.30 & 75.10 & 76.56 & 51.58 & 34.96 & 60.55 & 61.19 & 51.95\\
    
    \midrule
    \multicolumn{15}{l}{\textbf{BridgeAlign (Ours)}}\\
    \rowcolor[gray]{0.97}\hspace{4pt}+RubricPO & Qwen3-30B-A3B 
    & 49.93 & 68.44 & 65.81 & 68.84 & 63.54 
    & 57.22 & 75.07 & 76.94 & 51.32 & 35.30 & 60.61 & 61.93 & 51.80\\

    \rowcolor[gray]{0.97}\hspace{4pt}+BridgePO & Qwen3-30B-A3B
    & \textbf{51.45} & \textbf{72.20} & \textbf{66.90} & 70.06 & \textbf{65.35}
    & \textbf{58.96} & 74.93 & \textbf{77.33} & 51.55 & \textbf{36.24} & \textbf{61.17} & \textbf{63.05} & \textbf{53.80}\\

    \midrule
    \multicolumn{15}{l}{\textbf{BridgeAlign (Ablation Study)}}\\
    \hspace{4pt}+H-MPO & Qwen3-30B-A3B
    & 44.56 & 64.84 & 62.04 & 69.94 & 60.53
    & 55.27 & 75.06 & 76.88 & 51.57 & 35.29 & 60.29 & 60.40 & 50.18\\

    \hspace{4pt}+BridgePO v1 & Qwen3-30B-A3B
    & 50.25 & 68.90 & 65.16 & 67.10 & 63.11
    & 57.05 & 74.94 & 76.72 & 51.05 & 35.54 & 60.50 & 61.67 & 51.48\\

    \hspace{4pt}+BridgePO v2 & Qwen3-30B-A3B
    & 49.64 & 67.37 & 65.13 & 64.84 & 62.12
    & 57.37 & 75.07 & 76.83 & 51.31 & 35.34 & 60.63 & 61.30 & 52.08\\

    \hspace{4pt}+BridgePO v1+v2+RubricPO & Qwen3-30B-A3B
    & 49.89 & 69.69 & 64.82 & 67.75 & 63.23
    & 57.68 & 74.97 & 76.92 & 51.44 & 35.38 & 60.70 & 61.84 & 52.62\\
    
    \bottomrule
  \end{tabular}
  }
  
  \caption{
    Results of different methods across 17 benchmarks spanning nine core abilities, including the adopted data synthesis models and separate AlpacaEval 2 win rates. ``Avg HP'' and ``Avg KB'' denote the averages on human-preference and knowledge benchmarks, respectively, while “Avg” denotes the average over all 17 benchmarks.
  }
  \label{tab:main_result}
  
\end{table*}

\subsection{Main Results}
Table~\ref{tab:main_result} shows the results of all methods after training on Qwen3-8B across 17 benchmarks. Full results are reported in Appendix Table~\ref{tab:full_results}. Further analysis offers interesting insights:

\begin{itemize}
% \begin{itemize}%[leftmargin=0pt,label={},itemsep=-5pt,topsep=2pt]
\item {\emph{Fine-grained preference alignment is crucial.}
The BridgePO variant achieves the best performance \emph{on average} among the 11 strong baselines compared (ours ranks first on the most benchmarks), benefiting from its controllable quality degradation that enables finer-grained alignment.}

\item {\emph{Leading performance on both human-preference and knowledge-based tasks.}}
On the four human-preference tasks, BridgePO consistently outperforms the strongest baseline, with gains of +0.20 to +1.28 points and a +0.71 gain on Avg HP.
On the five knowledge-based tasks, it ranks first on Avg KB, with the largest gain on instruction following. More importantly, BridgePO offers a reusable pipeline for subjective long-form alignment, with consistent gains that are meaningful under strong baselines.

\item \emph{Data diversity and distillation are effective.} Among baselines, web-driven SynthQuestions and instruction-distilled Magpie perform well, indicating that both diverse data sources and well-constructed closed-source instruction tuning data are useful.
\item \emph{Hard rewards outperform soft constraints.} Under the same setting, Self-Rewarding outperforms FollowSoftConstraint.

\item \emph{Alpaca Eval Results.}
BridgePO achieves the highest AlpacaEval~2 win rate among all methods, followed by Self-Rewarding+M$_3$, while H-MPO performs the worst, suggesting that alignment optimization based on controlled document-quality degradation better supports instruction following. This is consistent with its leading Instruct score in Table~\ref{tab:main_result}, indicating that a richer instruction-following behavior translates into a higher win rate.

\end{itemize}

\begin{figure}[t]
    \centering
    \includegraphics[width=0.4\textwidth]{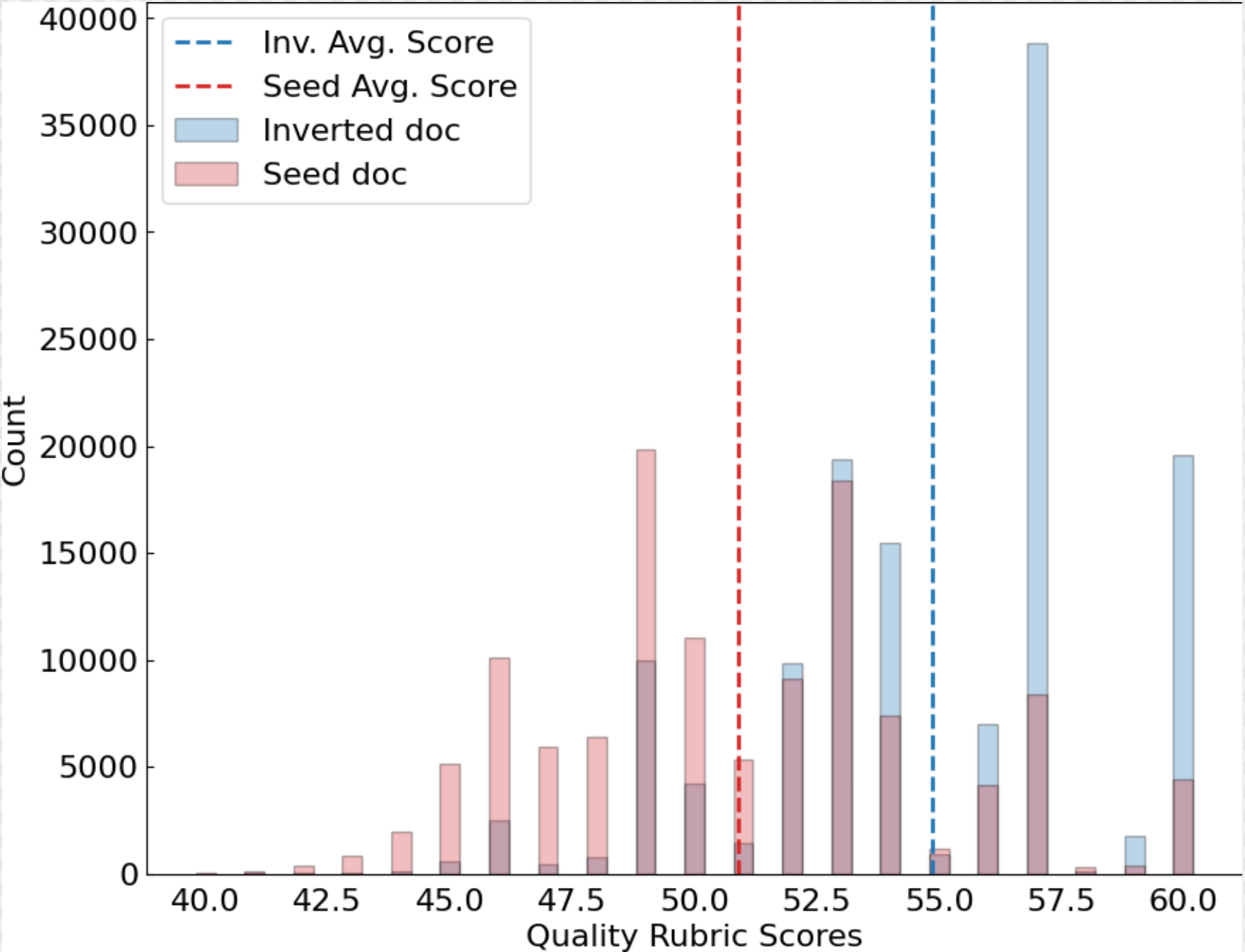}
    
    \caption{
    Distribution of HSS quality rubric scores for \textit{inverted} and \textit{seed} documents: 77.6\% of \textit{inverted} documents scored higher, 6.6\% equal, and 15.7\% lower than \textit{seed} documents.
    }
    \label{fig:inv_seed_score_dist}
    
\end{figure}

\begin{figure}[t]
    \centering
    \begin{subfigure}[b]{0.233\textwidth}
        \centering
        \includegraphics[width=\textwidth]{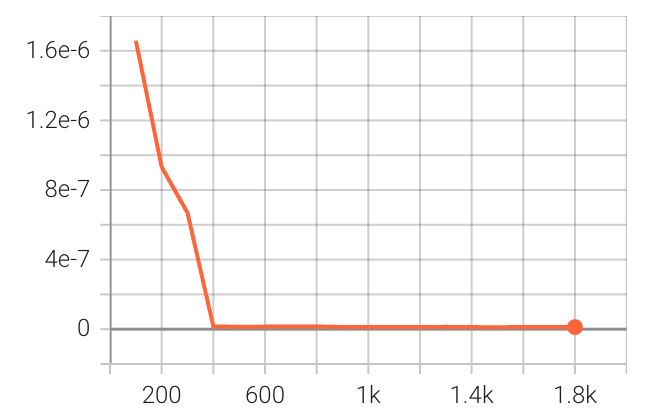}
        % \caption{H-MPO loss curve}
        \label{fig:hmpo_loss}
    \end{subfigure}
    \hfill
    \begin{subfigure}[b]{0.233\textwidth}
        \centering
        \includegraphics[width=\textwidth]{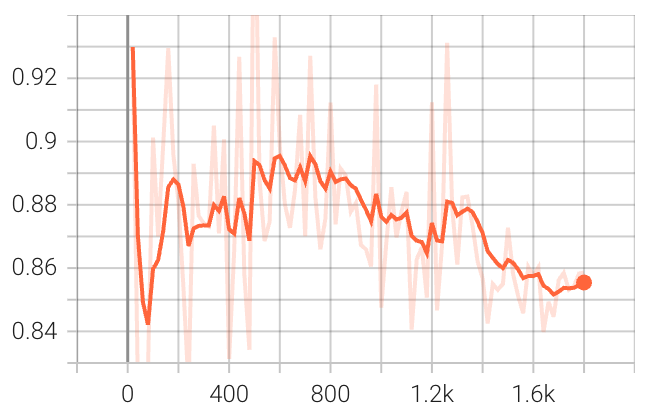}
        % \caption{RubricPO loss curve}
        \label{fig:rubricpo_loss}
    \end{subfigure}
    
    \caption{Convergence curves of training loss: H-MPO (left) vs. RubricPO (right)}
    \label{fig:loss_compare}
    
\end{figure}

\subsection{Ablation Studies}

\paragraph{RubricPO Outperforms H-MPO}~\label{RubricPO_vs_H-MPO} 
Table~\ref{tab:main_result} shows that RubricPO outperforms H-MPO, indicating that aligning with text HSS quality better satisfies human preferences and improves performance on standard instruction-following tasks, rather than on tasks like AlpacaEval~2 that favor concise, direct answers. 
Moreover, H-MPO relies solely on data source, treating web-sourced seed documents as chosen and LLM-generated inverted documents as rejected, which may easily mislabel quality: 
i) in fact, seed documents often have lower quality scores than inverted documents, as shown in Figure~\ref{fig:inv_seed_score_dist};
ii) human-model preferences may encourage the model to rely on superficial source cues rather than truth quality features, as evidenced by the faster training loss convergence of H-MPO compared to RubricPO in Figure~\ref{fig:loss_compare}.

\begin{figure*}[t]
    \centering
    \includegraphics[width=\textwidth]{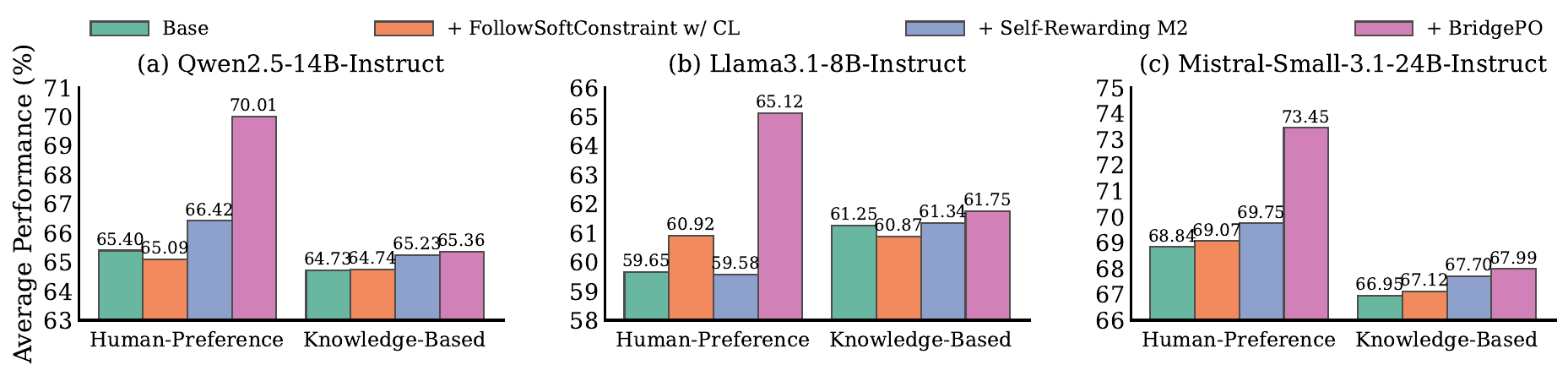}
    \caption{Performance comparison among the BridgeAlign BridgePO variant, the official instruction-tuned model, and two strong baselines across different model architectures and scales.}
    \label{fig:model_comparison}
    
\end{figure*}

\paragraph{Ablation on BridgePO Variants and Mixture with RubricPO}
Ablation study in Table~\ref{tab:main_result} also presents the results of two BridgePO variants and their mixture with RubricPO.
Among them, BridgePO-v1 only applies controllable quality simplification to inverted documents to generate transitional documents, and optimizes over the $\langle\text{inverted},\text{transitional}\rangle$ preference pairs.
BridgePO-v2 performs the same operation only on seed documents, optimizing over $\langle\text{seed},\text{transitional}\rangle$ pairs.
BridgePO v1+v2+RubricPO mixes these two sets of pairs with the rubric-based preference $\langle\text{inverted},\text{seed}\rangle$ and downsamples proportionally, covering all binary preferences among the three documents, i.e., full pairwise preference.
Ablation results show that all variants are inferior to the original BridgePO. This is because, although inverted documents usually score higher (77.6\%), seed documents surpass them in 15.7\% of cases, so selecting the higher-scoring document before degradation is more robust (see Figure~\ref{fig:inv_seed_score_dist}).

\begin{figure}[t]                         
  \centering
  \includegraphics[width=\linewidth]{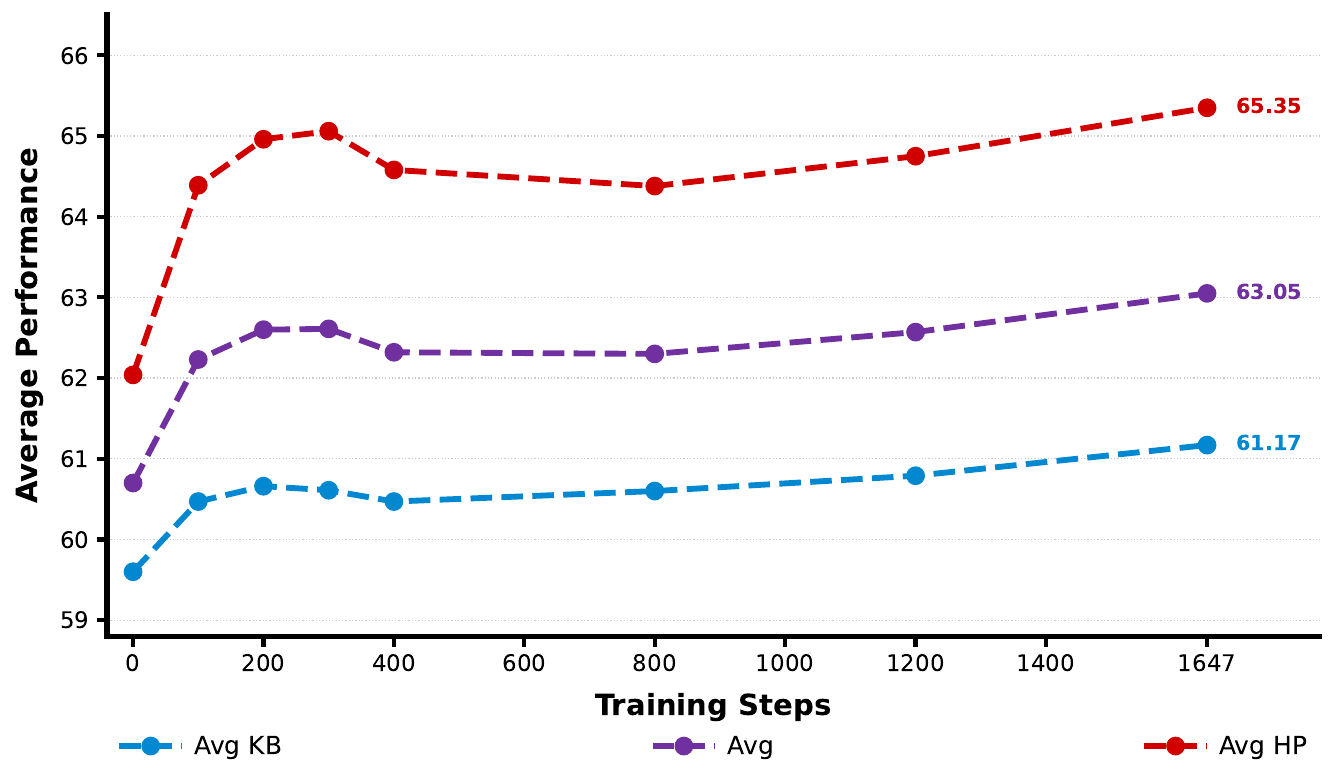}
  \caption{
Model performance versus data scale: the overall score rises rapidly before slowing after about 300 steps ($\approx$38.2k samples), with more pronounced gains in human-preference than knowledge-based abilities. Per-capability trends are shown in Appendix Figure~\ref{fig:scaling_effect_9panel}.
  }
  \label{fig:scaling_effect}
  
\end{figure}

\paragraph{Performance Across Model Architectures and Sizes}
Figure~\ref{fig:model_comparison} compares the official instruction-tuned models, our BridgeAlign variant BridgePO, and two strong baselines, Self-Rewarding-M2 and FollowSoftConstraint with curriculum learning, across Llama-3.1-8B, Qwen2.5-14B, and Mistral-Small-3.1-24B, representing variation in model architecture, scale, and both factors, respectively.
The results show that BridgePO consistently improves human-preference performance across all three models while maintaining slightly better results on knowledge-based benchmarks. This trend also shows that BridgeAlign generalizes well across model architectures and scales, making it a broadly applicable alignment framework for HSS domains.

\paragraph{Data Scaling Effect} 
Figure~\ref{fig:scaling_effect} shows the data scaling effect of BridgeAlign variant BridgePO, over one training epoch (210k preference samples, 1647 steps). The three average-score curves indicate that performance improves rapidly within the first 300 steps as more preference data are used, and then grows more slowly with diminishing marginal gains. Notably, the improvement in the human-preference average is more pronounced than that in the knowledge-based average.
Appendix Figure~\ref{fig:scaling_effect_9panel} further presents fine-grained trends across nine LLM capabilities.
\emph{Emotion perception, writing skill, and long context} peak early and then decline slightly, suggesting mild overfitting;
\emph{role-playing} shows only slight fluctuations after the initial gain;
\emph{reading comprehension and social interaction} rise early, dip in the middle, and rebound toward the end;
\emph{instruction following and commonsense reasoning} continue to benefit from later training, rising steadily toward the end, whereas \emph{world knowledge} stays largely flat with only minor fluctuations throughout.

\begin{table}[t]
\centering
% \small % removed: \resizebox controls final font size
% \renewcommand{\arraystretch}{1.0}
\setlength{\tabcolsep}{1pt}
\resizebox{1.0\columnwidth}{!}
{
\begin{tabular}{lccc}
\toprule
\textbf{Select Method} & \textbf{Avg. Input Len} & \textbf{Avg. Output Len} & \textbf{Lex-Div} \\
\midrule
Ultrafeedback & 159 & 281 & 63.10 \\
WritingPrompts & 53 & 666 & 88.49 \\
SynthQuestions & 121 & 717 & 70.00 \\
Magpie\_Air & 34 & 524 & 63.36 \\
Magpie\_Pro & 134 & 500 & 62.16 \\
Magpie\_Mixed & 83 & 491 & 71.19 \\
Self-Rewarding+M$_2$ & 263 & 1,547 & 93.87 \\
Self-Rewarding+M$_3$ & 153 & 1,145 & 87.19 \\
FollowSoftConstraint & 121 & 307 & 78.68 \\
\midrule
\textbf{BridgeAlign (ours)} &  &  &  \\
\rowcolor[gray]{0.97}\hspace{4pt}+RubricPO & 255 & 2,223 & 118.84 \\
\rowcolor[gray]{0.97}\hspace{4pt}+BridgePO & 254 & 2,193 & 118.59 \\
\bottomrule
\end{tabular}}

\caption{Average input/output token lengths and lexical diversity across datasets of different methods.}
\label{tab:lexical-diversity-rounded}

\end{table}

\section{In-Depth Analysis}

\paragraph{Basic Dataset Statistics}
{Table~\ref{tab:lexical-diversity-rounded} reports the basic statistics of datasets generated by different methods, including average input/output token lengths and lexical diversity (computed via the MTLD algorithm~\cite{mccarthy2010mtld}). Compared with non-HSS preference-data baselines, BridgeAlign has longer input and output sequences and higher lexical diversity, reaching 1.3 to 1.9 times that of other datasets. Controlled analyses related to token budget and HSS specificity are provided in Appendix~\ref{sec:controlled_ablations_appendix}.}

\begin{figure}[tb]
\centering
\begin{subfigure}{0.65\linewidth}
\centering
\includegraphics[width=\linewidth]{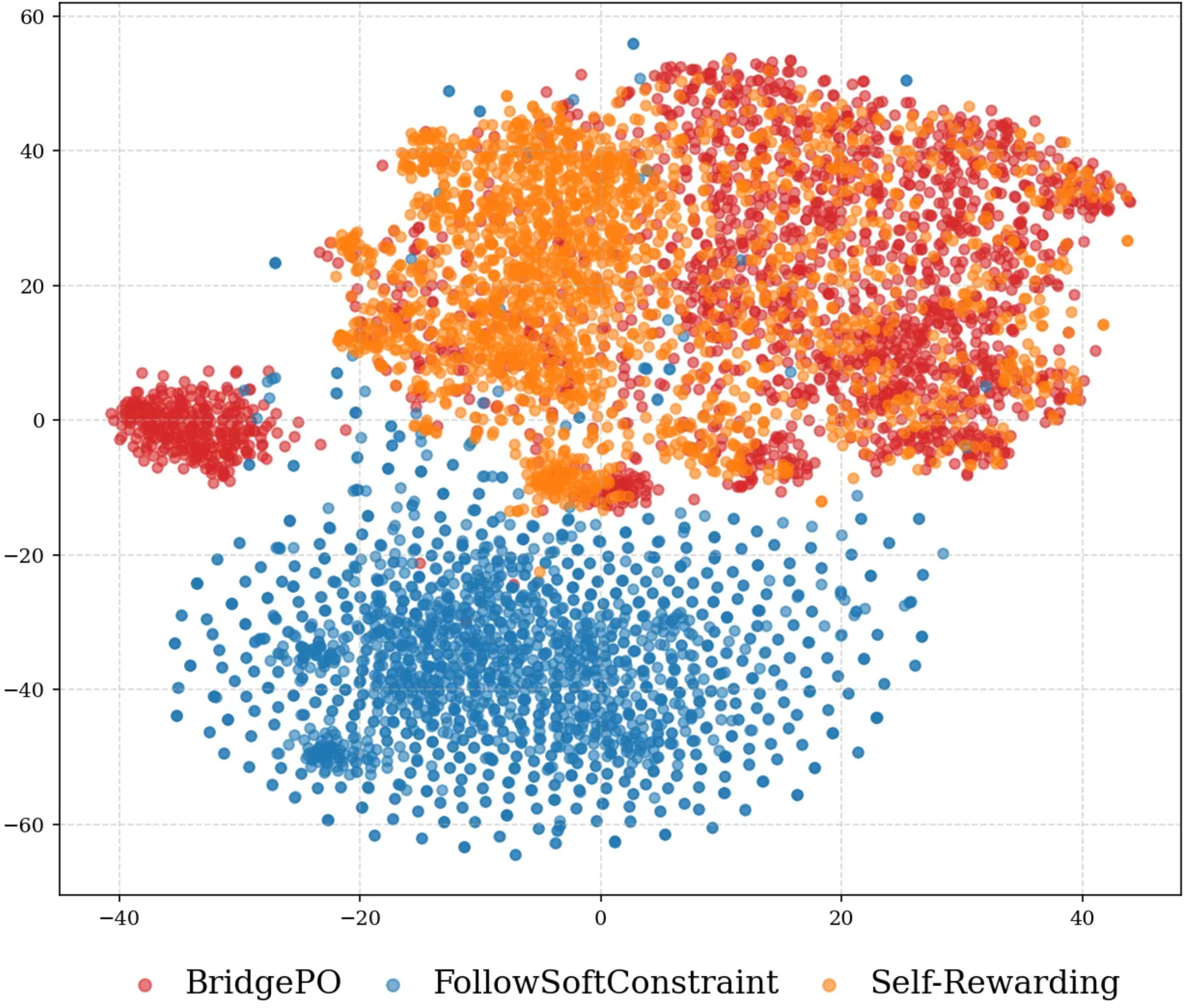}
\label{fig:tsne_embedding_alignment}
\end{subfigure}\hfill
\begin{subfigure}{0.35\linewidth}
\centering
\includegraphics[width=\linewidth]{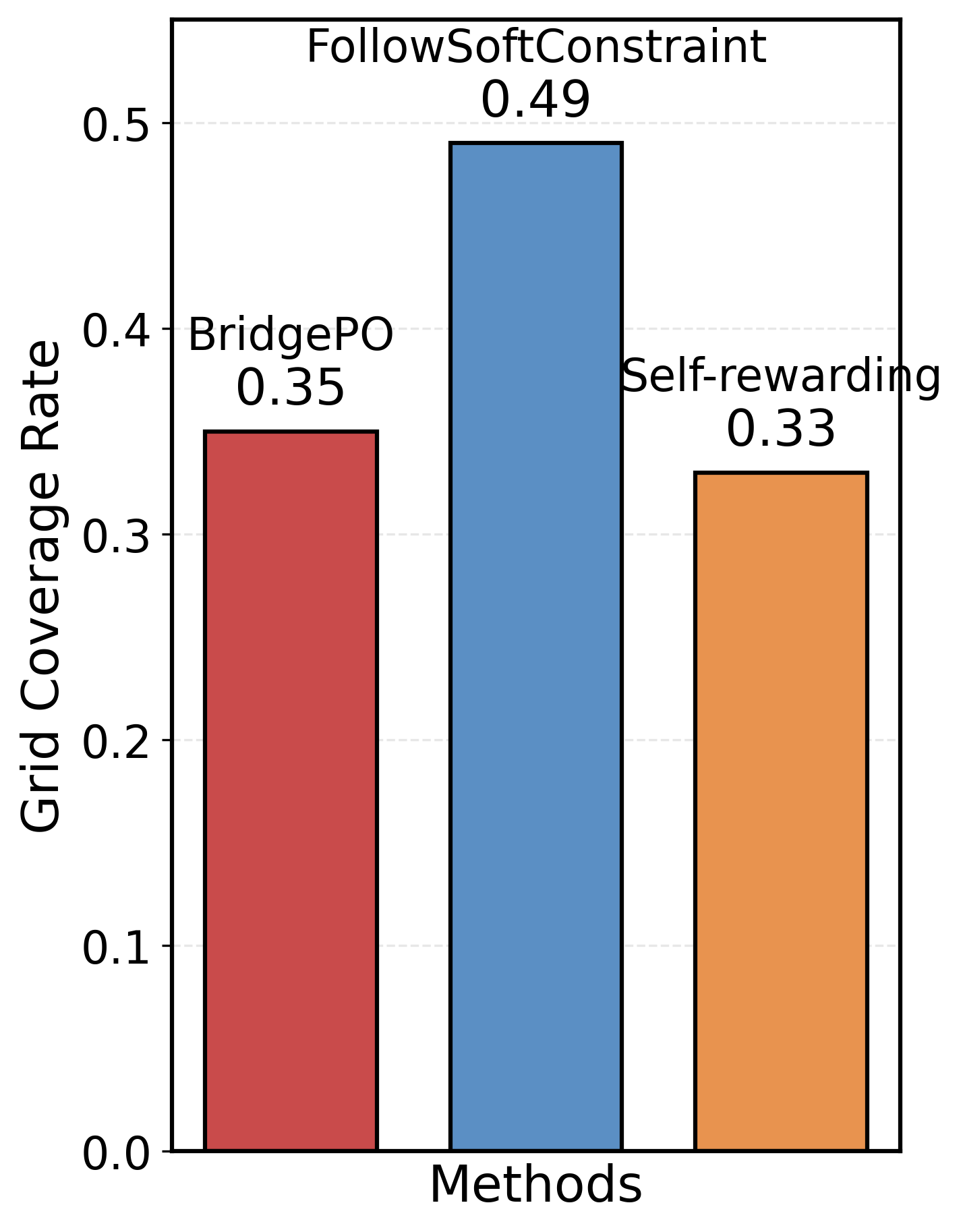}
\label{fig:grid_coverage_rate}
\end{subfigure}
\caption{t-SNE plot (left) and grid coverage rate (right) of sentence embeddings for BridgePO, FollowSoftConstraint, and Self-Rewarding.}
\label{fig:tsne_alignment}

\end{figure}

\paragraph{Data Semantic Diversity}
We visualize the semantic diversity of the data (i.e., instructions+preferred answers) generated by BridgeAlign (BridgePO) versus other baselines. 
We randomly sample 10k preference pairs from each dataset, compute their sentence embeddings using {gte-Qwen2-7B-instruct}~\cite{li2023towards}, and project them into a 2D space via t-SNE. Figure~\ref{fig:tsne_alignment} shows that the sample dispersity of BridgePO is comparable to that of the two baselines. 
Quantitative analysis further confirms that BridgePO ranks second in grid coverage, indicating broad semantic coverage that provides rich training signals and enhances generalization.

\begin{table}[tb]
\centering
\resizebox{1.0\columnwidth}{!}
{
\begin{tabular}{lcccc}
\toprule
             & \textbf{Avg HP} & \textbf{Avg KB} & \textbf{Avg} & \textbf{Alpaca Eval} \\
\midrule
Qwen3-8B     & 62.04 & 59.61 & 60.70    & 50.00      \\
Intruct-Tuning & 61.67 & 58.68 & 60.02 & 48.02      \\
\midrule
\textbf{BridgeAlign (Ours)}  &       &     &      &            \\
\hspace{4pt}+H-MPO       & 60.53 & 60.29 & 60.40 & 50.18 \\
\rowcolor[gray]{0.97}\hspace{4pt}+RubricPO      & 63.54 & 60.61 & 61.93 & 51.80      \\
\rowcolor[gray]{0.97}\hspace{4pt}+BridgePO      & 65.35 & 61.17 & 63.05 & 53.80      \\
\bottomrule
\end{tabular}
}

\caption{Performance comparison of preference alignment vs.\ instruction tuning. 
}\label{tab:sft_dpo_performance_comparison}
\vspace{-6pt}
\end{table}

\paragraph{Preference Alignment better suits HSS tasks than Instruction Tuning}
We built instruction–response pairs from inverted instructions and documents, fine-tuned Qwen3-8B on this dataset, and compared it against our preference alignment methods H-MPO, RubricPO, and BridgePO across all 17 human-preference and knowledge benchmarks in Table~\ref{tab:sft_dpo_performance_comparison}. Even the simplest PO method, H-MPO, still outperforms the instruction-tuned model on average, while other PO methods yield larger gains. 
These findings suggest that for open-domain HSS tasks, preference alignment is more effective than instruction tuning on fixed Q\&A pairs, as answers are rarely unique and differences lie mainly nuanced quality rather than objective correctness.

\begin{table}[tb]
\centering
% \small % removed: \resizebox controls final font size
% \renewcommand{\arraystretch}{1.0}
\setlength{\tabcolsep}{0.7mm}
\resizebox{1.0\columnwidth}{!}
{
\begin{tabular}{lcc}
\toprule
\textbf{Method} & \textbf{Human-touch} & \textbf{Rubric Score} \\
\midrule
Qwen3-8B & 17.56 & 56.21 \\
Self-Rewarding+M$_2$ & 17.53 & 56.17 \\ 
FollowSoftConstraint w/ CL & 17.55 & 56.19 \\
\midrule
\textbf{BridgeAlign (Ours)} &  &  \\ 
\rowcolor[gray]{0.97}\hspace{4pt} +RubricPO & 17.63 & 56.32 \\ 
\rowcolor[gray]{0.97}\hspace{4pt} +BridgePO & 17.82 & 56.63 \\
\bottomrule
\end{tabular}
}

\caption{Average human-touch metrics and rubric scores.}
\label{tab:human_touch}

\end{table}

\paragraph{Alignment on Textual HSS Quality}
We generate responses to 10k i.i.d. non-HSS queries using models trained by different methods, and use an LLM judge to evaluate their average human-touch metrics and HSS quality rubric scores. Table~\ref{tab:human_touch} shows that, even under this OOD setting, BridgeAlign outperforms the official Qwen3-8B and two non-HSS baselines on both scores, indicating that the preference-optimized model effectively learns HSS textual quality.

\begin{figure}[tb]
\centering
\includegraphics[width=0.9\columnwidth]{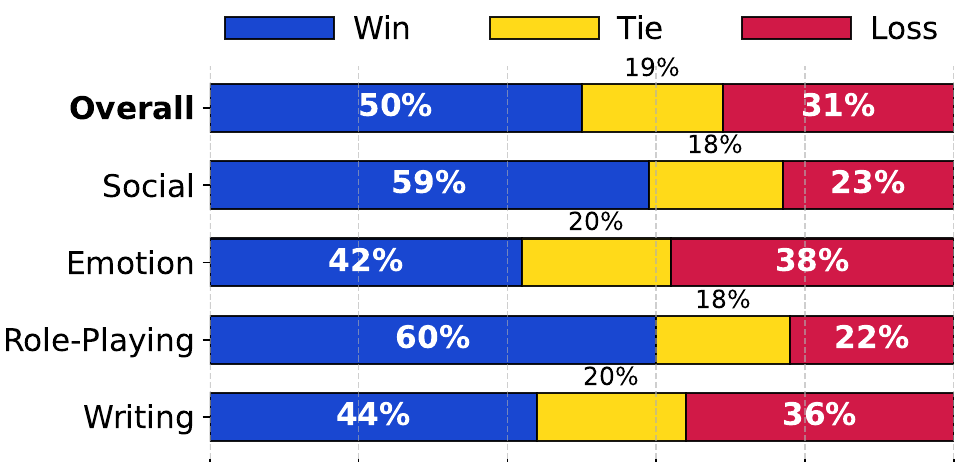}
\caption{Human win/tie/loss rates of BridgeAlign (+BridgePO) against Self-Rewarding+M$_3$ across tasks.}
\label{fig:human_eval_winrate}
\end{figure}

\paragraph{Human Evaluation}
To verify that our gains reflect real human preference rather than LLM-judge biases (e.g., length bias or model affinity), we conduct a blind pairwise human study on 120 prompts sampled from the human-preference benchmarks. For each prompt, three annotators compare the anonymized, randomly ordered responses of BridgeAlign (+BridgePO) against the strongest baseline (Self-Rewarding+M$_3$). As shown in Figure~\ref{fig:human_eval_winrate}, BridgePO attains a 50\% win rate (19\% tie, 31\% loss), with the largest margins on role-playing (60\%) and social interaction (59\%). This human ranking is consistent with the LLM-judge results in Table~\ref{tab:main_result}, confirming that the gains are real.

\vspace{-6pt}

\section{Conclusion}

In this paper, we propose BridgeAlign, a systematic HSS-tailored framework for preference data synthesis and alignment. Our three-stage pipeline—seed curation, preference data synthesis, and preference optimization—generates over 210k preference samples, applies multiple optimization algorithms on them, and achieves the best average performance on 17 benchmarks covering nine core LLM capabilities, leading on both human-preference and knowledge-based abilities simultaneously. 
Extensive experiments validate its effectiveness, while existing theory interprets its improvements.

% when the paper exceeds the allowable number of pages.
{\small
\bibliography{arxiv2026_bridgealign}

@inproceedings{peng2019neural,
  title={Neural machine translation with attention based on a new syntactic branch distance},
  author={Peng, Ru and Chen, Zhitao and Hao, Tianyong and Fang, Yi},
  booktitle={China Conference on Machine Translation},
  pages={47--57},
  year={2019},
  organization={Springer}
}

@inproceedings{liu2024best,
title={Best Practices and Lessons Learned on Synthetic Data},
author={Ruibo Liu and Jerry Wei and Fangyu Liu and Chenglei Si and Yanzhe Zhang and Jinmeng Rao and Steven Zheng and Daiyi Peng and Diyi Yang and Denny Zhou and Andrew M. Dai},
booktitle={First Conference on Language Modeling},
year={2024},
url={https://openreview.net/forum?id=OJaWBhh61C}
}

@inproceedings{
    yu2024metamath,
    title={MetaMath: Bootstrap Your Own Mathematical Questions for Large Language Models},
    author={Longhui Yu and Weisen Jiang and Han Shi and Jincheng YU and Zhengying Liu and Yu Zhang and James Kwok and Zhenguo Li and Adrian Weller and Weiyang Liu},
    booktitle={The Twelfth International Conference on Learning Representations},
    year={2024},
    url={https://openreview.net/forum?id=N8N0hgNDRt}
}

@article{austin2021program,
  title={Program synthesis with large language models},
  author={Austin, Jacob and Odena, Augustus and Nye, Maxwell and Bosma, Maarten and Michalewski, Henryk and Dohan, David and Jiang, Ellen and Cai, Carrie and Terry, Michael and Le, Quoc and others},
  journal={arXiv preprint arXiv:2108.07732},
  year={2021}
}

@article{cai2023large,
  title={Large language models as tool makers},
  author={Cai, Tianle and Wang, Xuezhi and Ma, Tengyu and Chen, Xinyun and Zhou, Denny},
  journal={arXiv preprint arXiv:2305.17126},
  year={2023}
}

@inproceedings{zhao2025tabula,
  title={Tabula: Harnessing language models for tabular data synthesis},
  author={Zhao, Zilong and Birke, Robert and Chen, Lydia Y},
  booktitle={Pacific-Asia Conference on Knowledge Discovery and Data Mining},
  pages={247--259},
  year={2025},
  organization={Springer}
}

@article{eldan2023tinystories,
  title={Tinystories: How small can language models be and still speak coherent english?},
  author={Eldan, Ronen and Li, Yuanzhi},
  journal={arXiv preprint arXiv:2305.07759},
  year={2023}
}

@article{yuan2024self,
  title={Self-rewarding language models},
  author={Yuan, Weizhe and Pang, Richard Yuanzhe and Cho, Kyunghyun and Sukhbaatar, Sainbayar and Xu, Jing and Weston, Jason},
  journal={arXiv preprint arXiv:2401.10020},
  volume={3},
  year={2024}
}

@article{wang2024weaver,
  title={Weaver: Foundation models for creative writing},
  author={Wang, Tiannan and Chen, Jiamin and Jia, Qingrui and Wang, Shuai and Fang, Ruoyu and Wang, Huilin and Gao, Zhaowei and Xie, Chunzhao and Xu, Chuou and Dai, Jihong and others},
  journal={arXiv preprint arXiv:2401.17268},
  year={2024}
}

@article{qian2025bottom,
  title={Bottom-Up Synthesis of Knowledge-Grounded Task-Oriented Dialogues with Iteratively Self-Refined Prompts},
  author={Qian, Kun and Chen, Maximillian and Li, Siyan and Sharma, Arpit and Yu, Zhou},
  journal={arXiv preprint arXiv:2504.14375},
  year={2025}
}

@article{yue2024mammoth2,
  title={MAmmoTH2: Scaling Instructions from the Web},
  author={Yue, Xiang and Zheng, Tuney and Zhang, Ge and Chen, Wenhu},
  journal={Advances in Neural Information Processing Systems},
  year={2024}
}

@article{ren2025step,
  title={Step-by-step mastery: Enhancing soft constraint following ability of large language models},
  author={Ren, Qingyu and Zeng, Jie and He, Qianyu and Liang, Jiaqing and Xiao, Yanghua and Zhou, Weikang and Sun, Zeye and Yu, Fei},
  journal={arXiv preprint arXiv:2501.04945},
  year={2025}
}

@article{gao2025principled,
  title={Principled data selection for alignment: The hidden risks of difficult examples},
  author={Gao, Chengqian and Li, Haonan and Liu, Liu and Xie, Zeke and Zhao, Peilin and Xu, Zhiqiang},
  journal={arXiv preprint arXiv:2502.09650},
  year={2025}
}

@article{wu2024beta,
  title={{\(\beta\)-DPO: Direct Preference Optimization with Dynamic \(\beta\)}},
  author={Wu, Junkang and Xie, Yuexiang and Yang, Zhengyi and Wu, Jiancan and Gao, Jinyang and Ding, Bolin and Wang, Xiang and He, Xiangnan},
  journal={Advances in Neural Information Processing Systems},
  volume={37},
  pages={129944--129966},
  year={2024}
}

@inproceedings{azar2024general,
  title={A general theoretical paradigm to understand learning from human preferences},
  author={Azar, Mohammad Gheshlaghi and Guo, Zhaohan Daniel and Piot, Bilal and Munos, Remi and Rowland, Mark and Valko, Michal and Calandriello, Daniele},
  booktitle={International Conference on Artificial Intelligence and Statistics},
  pages={4447--4455},
  year={2024},
  organization={PMLR}
}

@article{mchugh2012interrater,
  title={Interrater reliability: the kappa statistic},
  author={McHugh, Mary L},
  journal={Biochemia medica},
  volume={22},
  number={3},
  pages={276--282},
  year={2012},
  publisher={Hrvatsko dru{\v{s}}tvo za medicinsku biokemiju i laboratorijsku medicinu}
}

@misc{cerebras2023slimpajama,
    author = {Soboleva, Daria and Al-Khateeb, Faisal and Myers, Robert and Steeves, Jacob R and Hestness, Joel and Dey, Nolan},
    title = {{SlimPajama: A 627B token cleaned and deduplicated version of RedPajama}},
    year = 2023,
    howpublished = {\url{https://cerebras.ai/blog/slimpajama-a-627b-token-cleaned-and-deduplicated-version-of-redpajama}},
    url = {https://huggingface.co/datasets/cerebras/SlimPajama-627B},
}

@article{rae2021scaling,
  title={Scaling language models: Methods, analysis \& insights from training gopher},
  author={Rae, Jack W and Borgeaud, Sebastian and Cai, Trevor and Millican, Katie and Hoffmann, Jordan and Song, Francis and Aslanides, John and Henderson, Sarah and Ring, Roman and Young, Susannah and others},
  journal={arXiv preprint arXiv:2112.11446},
  year={2021}
}

@article{raffel2020exploring,
  title={Exploring the limits of transfer learning with a unified text-to-text transformer},
  author={Raffel, Colin and Shazeer, Noam and Roberts, Adam and Lee, Katherine and Narang, Sharan and Matena, Michael and Zhou, Yanqi and Li, Wei and Liu, Peter J},
  journal={Journal of machine learning research},
  volume={21},
  number={140},
  pages={1--67},
  year={2020}
}

@article{joulin2016fasttext,
  title={Fasttext. zip: Compressing text classification models},
  author={Joulin, Armand and Grave, Edouard and Bojanowski, Piotr and Douze, Matthijs and J{\'e}gou, H{\'e}rve and Mikolov, Tomas},
  journal={arXiv preprint arXiv:1612.03651},
  year={2016}
}

@article{penedo2024fineweb,
  title={The fineweb datasets: Decanting the web for the finest text data at scale},
  author={Penedo, Guilherme and Kydl{\'\i}{\v{c}}ek, Hynek and Lozhkov, Anton and Mitchell, Margaret and Raffel, Colin A and Von Werra, Leandro and Wolf, Thomas and others},
  journal={Advances in Neural Information Processing Systems},
  volume={37},
  pages={30811--30849},
  year={2024}
}

@inproceedings{broder1997resemblance,
  title={On the resemblance and containment of documents},
  author={Broder, Andrei Z},
  booktitle={Proceedings. Compression and Complexity of SEQUENCES 1997 (Cat. No. 97TB100171)},
  pages={21--29},
  year={1997},
  organization={IEEE}
}

@article{koksal2023longform,
  title={Longform: Effective instruction tuning with reverse instructions},
  author={K{\"o}ksal, Abdullatif and Schick, Timo and Korhonen, Anna and Sch{\"u}tze, Hinrich},
  journal={arXiv preprint arXiv:2304.08460},
  year={2023}
}

@inproceedings{
    li2024selfalignment,
    title={Self-Alignment with Instruction Backtranslation},
    author={Xian Li and Ping Yu and Chunting Zhou and Timo Schick and Omer Levy and Luke Zettlemoyer and Jason E Weston and Mike Lewis},
    booktitle={The Twelfth International Conference on Learning Representations},
    year={2024},
    url={https://openreview.net/forum?id=1oijHJBRsT}
}

@article{ge2024scaling,
  title={Scaling synthetic data creation with 1,000,000,000 personas},
  author={Ge, Tao and Chan, Xin and Wang, Xiaoyang and Yu, Dian and Mi, Haitao and Yu, Dong},
  journal={arXiv preprint arXiv:2406.20094},
  year={2024}
}

@article{li2024synthetic,
  title={Synthetic data (almost) from scratch: Generalized instruction tuning for language models},
  author={Li, Haoran and Dong, Qingxiu and Tang, Zhengyang and Wang, Chaojun and Zhang, Xingxing and Huang, Haoyang and Huang, Shaohan and Huang, Xiaolong and Huang, Zeqiang and Zhang, Dongdong and others},
  journal={arXiv preprint arXiv:2402.13064},
  year={2024}
}

@article{nayak2024learning,
  title={Learning to generate instruction tuning datasets for zero-shot task adaptation},
  author={Nayak, Nihal V and Nan, Yiyang and Trost, Avi and Bach, Stephen H},
  journal={arXiv preprint arXiv:2402.18334},
  year={2024}
}

@article{xu2024magpie,
  title={Magpie: Alignment Data Synthesis from Scratch by Prompting Aligned LLMs with Nothing},
  author={Xu, Zhangchen and Jiang, Fengqing and Niu, Luyao and Deng, Yuntian and Poovendran, Radha and Choi, Yejin and Lin, Bill Yuchen},
  journal={arXiv preprint arXiv:2406.08464},
  year={2024}
}

@article{li2023towards,
  title={Towards general text embeddings with multi-stage contrastive learning},
  author={Li, Zehan and Zhang, Xin and Zhang, Yanzhao and Long, Dingkun and Xie, Pengjun and Zhang, Meishan},
  journal={arXiv preprint arXiv:2308.03281},
  year={2023}
}

@article{lu2025mutual,
  title={Mutual Reinforcement of LLM Dialogue Synthesis and Summarization Capabilities for Few-Shot Dialogue Summarization},
  author={Lu, Yen-Ju and Hu, Ting-Yao and Koppula, Hema Swetha and Pouransari, Hadi and Chang, Jen-Hao Rick and Xia, Yin and Kong, Xiang and Zhu, Qi and Wang, Simon and Tuzel, Oncel and others},
  journal={arXiv preprint arXiv:2502.17328},
  year={2025}
}

@article{stiennon2020learning,
  title={Learning to summarize with human feedback},
  author={Stiennon, Nisan and Ouyang, Long and Wu, Jeffrey and Ziegler, Daniel and Lowe, Ryan and Voss, Chelsea and Radford, Alec and Amodei, Dario and Christiano, Paul F},
  journal={Advances in neural information processing systems},
  volume={33},
  pages={3008--3021},
  year={2020}
}

@article{mccarthy2010mtld,
  title={MTLD, vocd-D, and HD-D: A validation study of sophisticated approaches to lexical diversity assessment},
  author={McCarthy, Philip M and Jarvis, Scott},
  journal={Behavior research methods},
  volume={42},
  number={2},
  pages={381--392},
  year={2010},
  publisher={Springer}
}

@article{lee2025doctalk,
  title={DocTalk: Scalable Graph-based Dialogue Synthesis for Enhancing LLM Conversational Capabilities},
  author={Lee, Jing Yang and Bonab, Hamed and Zalmout, Nasser and Zeng, Ming and Lokegaonkar, Sanket and Lockard, Colin and Huang, Binxuan and Sarkhel, Ritesh and Wang, Haodong},
  journal={arXiv preprint arXiv:2507.05750},
  year={2025}
}

@article{li2025preference,
  title={Preference leakage: A contamination problem in llm-as-a-judge},
  author={Li, Dawei and Sun, Renliang and Huang, Yue and Zhong, Ming and Jiang, Bohan and Han, Jiawei and Zhang, Xiangliang and Wang, Wei and Liu, Huan},
  journal={arXiv preprint arXiv:2502.01534},
  year={2025}
}

@article{liu2024lipo,
  title={Lipo: Listwise preference optimization through learning-to-rank},
  author={Liu, Tianqi and Qin, Zhen and Wu, Junru and Shen, Jiaming and Khalman, Misha and Joshi, Rishabh and Zhao, Yao and Saleh, Mohammad and Baumgartner, Simon and Liu, Jialu and others},
  journal={arXiv preprint arXiv:2402.01878},
  year={2024}
}

@article{li20252d,
  title={2D-Curri-DPO: Two-Dimensional Curriculum Learning for Direct Preference Optimization},
  author={Li, Mengyang and Zhang, Zhong},
  journal={arXiv preprint arXiv:2504.07856},
  year={2025}
}

@article{fan2018hierarchical,
  title={Hierarchical neural story generation},
  author={Fan, Angela and Lewis, Mike and Dauphin, Yann},
  journal={arXiv preprint arXiv:1805.04833},
  year={2018}
}

@article{dubois2024length,
  title={Length-controlled alpacaeval: A simple way to debias automatic evaluators},
  author={Dubois, Yann and Galambosi, Bal{\'a}zs and Liang, Percy and Hashimoto, Tatsunori B},
  journal={arXiv preprint arXiv:2404.04475},
  year={2024}
}

@inproceedings{cui2023ultrafeedback,
  title={Ultrafeedback: Boosting language models with high-quality feedback.(2023)},
  author={Cui, Ganqu and Yuan, Lifan and Ding, Ning and Yao, Guanming and Zhu, Wei and Ni, Yuan and Xie, Guotong and Liu, Zhiyuan and Sun, Maosong},
  booktitle={URL https://openreview. net/forum},
  year={2023}
}

@article{zhu2025real,
  title={From Real to Synthetic: Synthesizing Millions of Diversified and Complicated User Instructions with Attributed Grounding},
  author={Zhu, Chiwei and Xu, Benfeng and Wang, Xiaorui and Mao, Zhendong},
  journal={arXiv preprint arXiv:2506.03968},
  year={2025}
}

@article{bai2024longwriter,
  title={Longwriter: Unleashing 10,000+ word generation from long context llms},
  author={Bai, Yushi and Zhang, Jiajie and Lv, Xin and Zheng, Linzhi and Zhu, Siqi and Hou, Lei and Dong, Yuxiao and Tang, Jie and Li, Juanzi},
  journal={arXiv preprint arXiv:2408.07055},
  year={2024}
}

@article{yang2025qwen3,
  title={Qwen3 technical report},
  author={Yang, An and Li, Anfeng and Yang, Baosong and Zhang, Beichen and Hui, Binyuan and Zheng, Bo and Yu, Bowen and Gao, Chang and Huang, Chengen and Lv, Chenxu and others},
  journal={arXiv preprint arXiv:2505.09388},
  year={2025}
}

@misc{qwen2.5,
    title = {Qwen2.5: A Party of Foundation Models},
    url = {https://qwenlm.github.io/blog/qwen2.5/},
    author = {{Qwen Team}},
    month = {September},
    year = {2024}
}

@article{paech2023eq,
  title={Eq-bench: An emotional intelligence benchmark for large language models},
  author={Paech, Samuel J},
  journal={arXiv preprint arXiv:2312.06281},
  year={2023}
}

@article{hendrycks2020measuring,
  title={Measuring massive multitask language understanding},
  author={Hendrycks, Dan and Burns, Collin and Basart, Steven and Zou, Andy and Mazeika, Mantas and Song, Dawn and Steinhardt, Jacob},
  journal={arXiv preprint arXiv:2009.03300},
  year={2020}
}

@article{mostafazadeh2016corpus,
  title={A corpus and evaluation framework for deeper understanding of commonsense stories},
  author={Mostafazadeh, Nasrin and Chambers, Nathanael and He, Xiaodong and Parikh, Devi and Batra, Dhruv and Vanderwende, Lucy and Kohli, Pushmeet and Allen, James},
  journal={arXiv preprint arXiv:1604.01696},
  year={2016}
}

@article{zhou2023instruction,
  title={Instruction-following evaluation for large language models},
  author={Zhou, Jeffrey and Lu, Tianjian and Mishra, Swaroop and Brahma, Siddhartha and Basu, Sujoy and Luan, Yi and Zhou, Denny and Hou, Le},
  journal={arXiv preprint arXiv:2311.07911},
  year={2023}
}

@article{yao2023collie,
  title={Collie: Systematic construction of constrained text generation tasks},
  author={Yao, Shunyu and Chen, Howard and Hanjie, Austin W and Yang, Runzhe and Narasimhan, Karthik},
  journal={arXiv preprint arXiv:2307.08689},
  year={2023}
}

@article{reddy2019coqa,
  title={Coqa: A conversational question answering challenge},
  author={Reddy, Siva and Chen, Danqi and Manning, Christopher D},
  journal={Transactions of the Association for Computational Linguistics},
  volume={7},
  pages={249--266},
  year={2019},
  publisher={MIT Press One Rogers Street, Cambridge, MA 02142-1209, USA journals-info~…}
}

@misc{eqbench3_repo_2025,
  author       = {Samuel J. Paech},
  title        = {EQ-Bench 3: Emotional Intelligence Benchmark},
  year         = {2025},
  howpublished = {\url{https://github.com/EQ-bench/eqbench3}},
  note         = {Commit \texttt{<hash>} or release \texttt{<tag>}
  }
}

@article{wu2025writingbench,
  title={Writingbench: A comprehensive benchmark for generative writing},
  author={Wu, Yuning and Mei, Jiahao and Yan, Ming and Li, Chenliang and Lai, Shaopeng and Ren, Yuran and Wang, Zijia and Zhang, Ji and Wu, Mengyue and Jin, Qin and others},
  journal={arXiv preprint arXiv:2503.05244},
  year={2025}
}

@article{zellers2019hellaswag,
  title={Hellaswag: Can a machine really finish your sentence?},
  author={Zellers, Rowan and Holtzman, Ari and Bisk, Yonatan and Farhadi, Ali and Choi, Yejin},
  journal={arXiv preprint arXiv:1905.07830},
  year={2019}
}

@article{shaham2022scrolls,
  title={Scrolls: Standardized comparison over long language sequences},
  author={Shaham, Uri and Segal, Elad and Ivgi, Maor and Efrat, Avia and Yoran, Ori and Haviv, Adi and Gupta, Ankit and Xiong, Wenhan and Geva, Mor and Berant, Jonathan and others},
  journal={arXiv preprint arXiv:2201.03533},
  year={2022}
}

@article{sap2019socialiqa,
  title={Socialiqa: Commonsense reasoning about social interactions},
  author={Sap, Maarten and Rashkin, Hannah and Chen, Derek and LeBras, Ronan and Choi, Yejin},
  journal={arXiv preprint arXiv:1904.09728},
  year={2019}
}

@article{narayan2018don,
  title={Don't give me the details, just the summary! topic-aware convolutional neural networks for extreme summarization},
  author={Narayan, Shashi and Cohen, Shay B and Lapata, Mirella},
  journal={arXiv preprint arXiv:1808.08745},
  year={2018}
}

@article{kovcisky2018narrativeqa,
  title={The narrativeqa reading comprehension challenge},
  author={Ko{\v{c}}isk{\`y}, Tom{\'a}{\v{s}} and Schwarz, Jonathan and Blunsom, Phil and Dyer, Chris and Hermann, Karl Moritz and Melis, G{\'a}bor and Grefenstette, Edward},
  journal={Transactions of the Association for Computational Linguistics},
  volume={6},
  pages={317--328},
  year={2018},
  publisher={MIT Press One Rogers Street, Cambridge, MA 02142-1209, USA journals-info~…}
}

@article{wang2023rolellm,
  title={Rolellm: Benchmarking, eliciting, and enhancing role-playing abilities of large language models},
  author={Wang, Zekun Moore and Peng, Zhongyuan and Que, Haoran and Liu, Jiaheng and Zhou, Wangchunshu and Wu, Yuhan and Guo, Hongcheng and Gan, Ruitong and Ni, Zehao and Yang, Jian and others},
  journal={arXiv preprint arXiv:2310.00746},
  year={2023}
}

@article{chen2024tombench,
  title={Tombench: Benchmarking theory of mind in large language models},
  author={Chen, Zhuang and Wu, Jincenzi and Zhou, Jinfeng and Wen, Bosi and Bi, Guanqun and Jiang, Gongyao and Cao, Yaru and Hu, Mengting and Lai, Yunghwei and Xiong, Zexuan and others},
  journal={arXiv preprint arXiv:2402.15052},
  year={2024}
}

@misc{vonwerra2022trl,
  author = {Leandro von Werra and Younes Belkada and Lewis Tunstall and Edward Beeching and Tristan Thrush and Nathan Lambert and Shengyi Huang and Kashif Rasul and Quentin Gallouédec},
  title = {TRL: Transformer Reinforcement Learning},
  year = {2020},
  publisher = {GitHub},
  journal = {GitHub repository},
  howpublished = {\url{https://github.com/huggingface/trl}}
}

@article{ouyang2022training,
  title={Training language models to follow instructions with human feedback},
  author={Ouyang, Long and Wu, Jeffrey and Jiang, Xu and Almeida, Diogo and Wainwright, Carroll and Mishkin, Pamela and Zhang, Chong and Agarwal, Sandhini and Slama, Katarina and Ray, Alex and others},
  journal={Advances in neural information processing systems},
  volume={35},
  pages={27730--27744},
  year={2022}
}

@article{rafailov2024direct,
  title={Direct preference optimization: Your language model is secretly a reward model},
  author={Rafailov, Rafael and Sharma, Archit and Mitchell, Eric and Manning, Christopher D and Ermon, Stefano and Finn, Chelsea},
  journal={Advances in Neural Information Processing Systems},
  volume={36},
  year={2024}
}

@article{dubey2024llama,
  title={The llama 3 herd of models},
  author={Dubey, Abhimanyu and Jauhri, Abhinav and Pandey, Abhinav and Kadian, Abhishek and Al-Dahle, Ahmad and Letman, Aiesha and Mathur, Akhil and Schelten, Alan and Yang, Amy and Fan, Angela and others},
  journal={arXiv preprint arXiv:2407.21783},
  year={2024}
}

@article{zheng2023judging,
  title={Judging llm-as-a-judge with mt-bench and chatbot arena},
  author={Zheng, Lianmin and Chiang, Wei-Lin and Sheng, Ying and Zhuang, Siyuan and Wu, Zhanghao and Zhuang, Yonghao and Lin, Zi and Li, Zhuohan and Li, Dacheng and Xing, Eric and others},
  journal={Advances in Neural Information Processing Systems},
  volume={36},
  pages={46595--46623},
  year={2023}
}

@inproceedings{wang2024codeclm,
  title={Codeclm: Aligning language models with tailored synthetic data},
  author={Wang, Zifeng and Li, Chun-Liang and Perot, Vincent and Le, Long and Miao, Jin and Zhang, Zizhao and Lee, Chen-Yu and Pfister, Tomas},
  booktitle={Findings of the Association for Computational Linguistics: NAACL 2024},
  pages={3712--3729},
  year={2024}
}

@inproceedings{gan2025towards,
  title={Towards a theoretical understanding of synthetic data in llm post-training: A reverse-bottleneck perspective},
  author={Gan, Zeyu and Liu, Yong},
  booktitle={International Conference on Learning Representations},
  volume={2025},
  pages={11212--11235},
  year={2025}
}

@article{bai2022constitutional,
  title={Constitutional ai: Harmlessness from ai feedback},
  author={Bai, Yuntao and Kadavath, Saurav and Kundu, Sandipan and Askell, Amanda and Kernion, Jackson and Jones, Andy and Chen, Anna and Goldie, Anna and Mirhoseini, Azalia and McKinnon, Cameron and others},
  journal={arXiv preprint arXiv:2212.08073},
  year={2022}
}

@article{lee2023rlaif,
  title={Rlaif vs. rlhf: Scaling reinforcement learning from human feedback with ai feedback},
  author={Lee, Harrison and Phatale, Samrat and Mansoor, Hassan and Mesnard, Thomas and Ferret, Johan and Lu, Kellie and Bishop, Colton and Hall, Ethan and Carbune, Victor and Rastogi, Abhinav and others},
  journal={arXiv preprint arXiv:2309.00267},
  year={2023}
}

@article{tunstall2023zephyr,
  title={Zephyr: Direct distillation of lm alignment},
  author={Tunstall, Lewis and Beeching, Edward and Lambert, Nathan and Rajani, Nazneen and Rasul, Kashif and Belkada, Younes and Huang, Shengyi and Von Werra, Leandro and Fourrier, Cl{\'e}mentine and Habib, Nathan and others},
  journal={arXiv preprint arXiv:2310.16944},
  year={2023}
}
}

% this arXiv build is not bound by that restriction.)
\clearpage


% ============================================================
% ================= TECHNICAL APPENDIX =======================
% (Merged single-file build for arXiv. Section letters A,B,...
%  and continued figure/table numbering follow from \appendix.)
% ============================================================
\appendix

\begin{table*}[t]
  \centering
  \small
  \renewcommand{\arraystretch}{0.9}
  \setlength{\tabcolsep}{3pt}
  \resizebox{1.0\textwidth}{!}{
  \begin{tabular}{l c c c c c}
    \toprule
    Score Range &
    Min.\ Sub-scores (A/B/C) &
    Quality Level &
    \#Docs &
    \#Tokens &
    Avg.\ Len \\
    \midrule
    55--60 &
    A$\ge$5, B$\ge$5, C$\ge$4 &
    Near-perfect &
    294,493 &
    1,103,904,718 &
    3,748 \\
    \addlinespace
    \textbf{45--54} &
    \textbf{A$\ge$5, B$\ge$4, C$\ge$3} &
    \textbf{Seed quality bar} &
    \textbf{1,073,755} &
    \textbf{3,153,687,402} &
    \textbf{2,939} \\
    \addlinespace
    31--44 &
    A$\ge$3, B$\ge$3, C$\ge$2 &
    Barely acceptable &
    12,137,698 &
    12,526,930,657 &
    1,032 \\
    \addlinespace
    $\le$30 or any A$\le$2 &
    -- &
    Fails readability/applicability &
    1,495,491 &
    1,621,533,520 &
    1,084 \\
    \bottomrule
  \end{tabular}}
  
  \caption{
  Corpus statistics grouped by HSS quality score.
  Score Range denotes the total-score interval.
  Min.\ Sub-scores specifies the minimum required scores for each quality rubric within tiers A (Readability), B (Applicability), and C (Human-touch).
  \#Docs and \#Tokens indicate corpus size, measured using the Qwen3 tokenizer.
  Avg.\ Len is the average document length in tokens.
  }
  \label{tab:quality_bands}
  
\end{table*}

\section{Details of HSS Quality Rubrics}\label{sec:hss_quality_rubrics}
Under expert guidance, we design 12 quality rubrics for evaluating HSS texts. Each rubric is scored on a 1–5 scale and organized into three hierarchically weighted dimensions:
\begin{enumerate}[label=\Alph*.]
% \begin{enumerate}[topsep=2pt, itemsep=1pt, parsep=0pt, leftmargin=*, label=\Alph*.]
    \item {Readability}: Content Accuracy, Coherence, Grammar, Domain Relevance (weight 0.5 × 4, max 10);
    \item {Applicability}: Vocabulary Richness, Knowledge Depth, Tone\& Expression, Gene Focus (weight 1.0 × 4, max 20);
    \item {Human-touch}: Literary Diversity, Emotionality, Thematic Depth, Humanities Creativity (weight 1.5 × 4, max 30).
\end{enumerate}
The total score ranges from 0 to 60. Based on the total score and minimum per-dimension requirements, we group the corpus into four quality bands (Table~\ref{tab:quality_bands}). Setting the seed threshold to 45 ensures perfect readability (A$\geq$5), solid applicability (B$\geq$4), and moderate human-touch (C$\geq$3), striking an optimal balance between data quality and scale. Raising the threshold further yields only marginal per-document gains while sharply reducing corpus size and long-tail coverage, thereby harming downstream instruction synthesis.

\section{Domain Distribution}

Table~\ref{tab:domain_composition} shows the domain distribution of the 210k-sample BridgeAlign preference alignment dataset. The dataset covers 14 target HSS domains, providing broad disciplinary coverage but also showing a clear imbalance. We believe this pattern mainly reflects how much web text is available in different HSS fields, rather than a bias in data collection: text-heavy domains such as History and Literature naturally provide more text data~\cite{peng2019neural}, while Geography and Arts depend more on non-text materials such as maps and images, and Healthcare is more limited by privacy regulations. Since manually balancing the data may cause overfitting in low-resource domains, we preserve the natural domain distribution and instead rely on high-quality synthesis to mitigate data quantity disparities.

\begin{table}[t]
\centering
\small
\setlength{\tabcolsep}{6pt}
\renewcommand{\arraystretch}{1.0}
\begin{tabular}{lrr}
\toprule
\textbf{Domain} & \textbf{Documents} & \textbf{Proportion (\%)} \\
\midrule
History & 49,680 & 23.66 \\
Arts & 31,370 & 14.94 \\
Politics & 25,588 & 12.18 \\
Literature & 24,546 & 11.69 \\
Philosophy & 13,425 & 6.39 \\
Sociology & 11,706 & 5.57 \\
Economics & 11,589 & 5.52 \\
Law & 8,173 & 3.89 \\
Sports & 7,548 & 3.59 \\
Psychology & 7,266 & 3.46 \\
Education & 6,773 & 3.23 \\
Management & 6,763 & 3.22 \\
Healthcare & 4,529 & 2.16 \\
Geography & 1,044 & 0.50 \\
\midrule
Total & 210,000 & 100.00 \\
\bottomrule
\end{tabular}

\caption{Domain distribution of the BridgeAlign dataset. Documents and proportions denote the number of samples and the proportion within each domain, respectively.}
\label{tab:domain_composition}

\end{table}

\section{Training Details}\label{sec:training_details}

Our training is conducted on the Huggingface TRL\footnote{\url{https://github.com/huggingface/trl}} ~\cite{vonwerra2022trl} using $8\times8$ NVIDIA H800 GPUs. One epoch for training the {Qwen3-8B} model was approximately 5 hours. We used the last saved checkpoint as the final model. Following the Qwen3 and Llama3.1 technical reports~\cite{yang2025qwen3,dubey2024llama}, the detailed training hyperparameters for H-MPO, RubricPO, and BridgePO are listed in Table~\ref{tab:dpo-hparams}.

\begin{table}[t]
\centering
\renewcommand{\arraystretch}{0.85}
\setlength{\tabcolsep}{10pt}
% \resizebox{1.0\columnwidth}{!}
    {
    \begin{tabular}{@{}ll@{}}
        \toprule
        Hyper-parameter            & Value \\ \midrule
        Epochs                     & 1 \\
        Learning rate              & $7\times10^{-7}$ \\
        Batch size                 & 128 \\
        Max length                 & 5120 \\
        Scheduler                  & Cosine \\
        Warm-up ratio              & 0.03 \\
        Weight decay               & 0.0 \\
        Adam $\beta_{2}$           & 0.999 \\
        $\beta$                    & 0.1 \\
        Gradient accumulation      & 1 \\
        Gradient checkpointing     & True \\
        Precision                  & BF16 \\ \bottomrule
    \end{tabular}}
    
    \caption{Training hyper-parameters of H-MPO, RubricPO, BridgePO.}
    \label{tab:dpo-hparams}

\end{table}

\section{Benchmarks and Metrics Details}\label{sec:benchmarks}
We adopt \textbf{17 mainstream benchmarks} spanning \textbf{nine core LLM capabilities}, grouped into two principal classes: human preference and knowledge-based benchmarks. 
For each benchmark, we report the sample size/setting (few-/zero-shot), evaluation metrics, and the judge model used when applicable; the absence of a listed judge indicates that no external judge is used. Thus, the benchmark suite covers both judge-based and objective evaluations, and incorporates multiple repeated runs for the judge-based subjective benchmarks (the number of runs is annotated per benchmark below, and reported scores are averaged over these runs) as a pragmatic evaluation compromise under current practical constraints. We further complement these winrate metrics with a blind pairwise human evaluation (detailed below).
The 17 benchmarks yield 20 metrics in total, since RoleBench contributes two metrics (IG, RG) and MMLU contributes three metrics (MMLU-Full, MMLU-Humanities, MMLU-Social-Sciences). Accordingly, the \textbf{Avg} column is computed over these 20 metrics, while AlpacaEval 2 is reported separately.

\begin{enumerate}%[topsep=2pt, itemsep=1pt, parsep=0pt, leftmargin=*]
    \item {Human Preference Benchmarks}
    \begin{itemize}%[topsep=2pt, itemsep=1pt, parsep=0pt, leftmargin=-2pt]
        \item {Emotion Perception:} {BuzzBench}~\cite{eqbench3_repo_2025} (zero-shot, rubric score, {Claude-3.5-Sonnet-v2}, avg. of 10 runs), {EQ-Bench3}~\cite{eqbench3_repo_2025} (zero-shot, rubric Score, {Claude-3.7-Sonnet}, avg. of 10 runs).

        \item {Role-Playing:} {RoleBench\_IG\_En}, {RoleBench\_RG\_En}~\cite{wang2023rolellm} (zero-shot, win rate; {Llama-3.3-70B-Instruct}, avg. of 3 runs). Since the built-in ROUGE-L metric is skewed by hallucinated ground truths, we adopt the Alpaca-Eval setup\footnote{\url{https://github.com/tatsu-lab/alpaca_eval}}, letting Llama-3.3-70B-Instruct pairwise compare answers with the ground truth and report the win rate.
        
        \item {Writing Skill:} {CreativeWriting-v3}~\cite{paech2023eq} (zero-shot, rubric score, {Claude-3.7-Sonnet}, avg. of 3 runs), {WritingBench}~\cite{wu2025writingbench} (zero-shot, rubric score, {Claude-3.7-Sonnet}, avg. of 3 runs), {Judgemark-v2}~\cite{paech2023eq} (zero-shot, judgemark Score, avg. of 20 runs).
        
        \item {Social Interactions:} {Social-IQA}~\cite{sap2019socialiqa} (zero-shot, accuracy), {IQuiz\_EQ}~\cite{chen2024tombench} (zero-shot, accuracy).
    \end{itemize}

    \item {Knowledge-Based Benchmarks}
    \begin{itemize}%[topsep=2pt, itemsep=1pt, parsep=0pt, leftmargin=-2pt]
        \item {Instruction Following:} {IFEval}~\cite{zhou2023instruction} (zero-shot, prompt-level strict accuracy), {Collie}~\cite{yao2023collie} (zero-shot, aggregated accuracy).
        
        \item {World Knowledge:} {MMLU}, {MMLU\_Humanities}, {MMLU\_Social\_Sciences}~\cite{hendrycks2020measuring} (5-shot, accuracy).
        
        \item {Commonsense Reasoning:} {HellaSwag}~\cite{zellers2019hellaswag} (6-shot, normalized accuracy), {StoryCloze}~\cite{mostafazadeh2016corpus} (zero-shot, accuracy).
        
        \item {Long Context:} {CoQA}~\cite{reddy2019coqa} (zero-shot, F1), {GovernmentReport\_CRS}~\cite{shaham2022scrolls} (zero-shot, ROUGE-L).
        
        \item {Reading Comprehension:} {NarrativeQA summarizeOnly}~\cite{kovcisky2018narrativeqa} (zero-shot, accuracy, {Llama-3.3-70B-Instruct}, avg. of 3 runs), {XSum}~\cite{narayan2018don} (zero-shot, ROUGE-L).
    \end{itemize}

    \item {Human Evaluation Benchmark}

        We construct a blind pairwise winrate benchmark to validate the performance gains beyond LLM-based judges. Specifically, we sample 120 prompts from the human-preference benchmarks (40 writing, 30 role-playing, 30 emotion, and 20 social interaction), and compare BridgeAlign (+BridgePO) against the strongest baseline (Self-Rewarding+M$_3$) in a pairwise manner. For each prompt, the two responses are presented side by side with their order shuffled and sources hidden, so that annotators cannot identify which model produced each response. Three annotators independently label each pair as a \emph{win}, \emph{tie}, or \emph{loss} for BridgePO based on the overall response quality, following the same evaluation metric used for the LLM judge. We report the majority-voted win/tie/loss rates per category and overall in Figure~\ref{fig:human_eval_winrate}.
\end{enumerate}

\begin{table*}[t]
\centering
\small
\renewcommand{\arraystretch}{1.0}
\resizebox{1.\textwidth}{!}{
\begin{tabular}{llccc}
\toprule
\textbf{LLM Capability} & \textbf{Benchmark Dataset} & \textbf{Unique 13-gram Number} & \textbf{13-gram Overlap Ratio} & \textbf{MinHash Near-duplicates} \\
\midrule
\textbf{Emotion} & BuzzBench & 5,382 & $0.00\%$ & $0.00\%$ \\
\textbf{Role} & RoleBench\_IG\_En & 12,851 & $0.00\%$ & $0.00\%$ \\
\textbf{Writing} & Creative Writing & 5,819 & $0.00\%$ & $0.00\%$ \\
\textbf{Social} & SIQA & 279,091 & $0.00\%$ & $0.00\%$ \\
\textbf{Instruction} & IFEval & 13,299 & $0.00\%$ & $0.00\%$ \\
\textbf{Knowledge} & MMLU & 398,281 & $0.00\%$ & $0.00\%$ \\
\textbf{Common} & HellaSwag & 1,262,209 & $0.00\%$ & $0.00\%$ \\
\textbf{Long} & CoQA & 2,325,293 & $0.00\%$ & $0.00\%$ \\
\textbf{Read} & NarrativeQA & 4,468,262 & $0.00\%$ & $0.00\%$ \\
\bottomrule
\end{tabular}}

\caption{Detailed decontamination analyses over nine LLM capabilities demonstrate that BridgeAlign has zero overlap with the evaluation benchmarks.}
\label{tab:decontamination_full}
\end{table*}

\section{Data Leakage Analysis}~\label{sec:appendix_data_decontamination}
To avoid evaluation leakage, we conduct a strict data decontamination analysis by selecting one representative benchmark from each of the nine core LLM capabilities. Following standard practices in recent technical reports~\citep{qwen2.5,dubey2024llama}, we use both \textbf{13-gram exact matching} and \textbf{MinHash LSH}, with a Jaccard threshold of $0.7$ and $128$ hash functions, to detect overlap between the queries in our synthesized BridgeAlign dataset and the queries in the evaluation benchmarks. As shown in Table~\ref{tab:decontamination_full}, both the exact-overlap ratio and the near-duplicate rate are \emph{zero} ($0.00\%$) for all benchmarks, indicating that BridgeAlign does not overlap with the evaluation sets and thus preserves evaluation fairness and integrity.

\section{Full results}
Table \ref{tab:full_results} presents the full results across 17 mainstream benchmarks covering nine core capabilities (20 metrics in total), with AlpacaEval 2 reported separately. Qwen3-8B aligned on BridgeAlign synthetic data achieves a balance between human preference and knowledge-base tasks, with the best average across all compared methods. On AlpacaEval~2, BridgePO attains the highest win rate among all methods.

\begin{table*}[t]
\centering
\scriptsize
\setlength{\tabcolsep}{1.2pt}
\renewcommand{\arraystretch}{1.1}
\resizebox{1.0\textwidth}{!}{
%--------------------------------------------------------
\begin{tabular}{
l
*{2}{c}  % Emotion (Buzz, EQ3)
*{2}{c}  % Role (IG, RG)
*{3}{c}  % Writing (CW, WB, JM)
*{2}{c}  % Social (SIQA, IQuiz)
*{2}{c}  % Instruct
*{3}{c}  % World
*{2}{c}  % Common
*{2}{c}  % Long
*{2}{c}  % Read
c        % Avg
c        % Alpaca v2
}
\toprule
\multirow{3}{*}{\textbf{Selection Method}}
& \multicolumn{9}{c}{\textbf{Human Preference}}
& \multicolumn{11}{c}{\textbf{Knowledge Based}}
& \multirow{3}{*}{\textbf{Avg}}
& \multirow{3}{*}{\textbf{AlpacaEval2}} \\

\cmidrule(lr){2-10}\cmidrule(lr){11-21}

%----------- 2nd‐level heads (Reordered) -----------
& \multicolumn{2}{c}{\textbf{Emotion}}
& \multicolumn{2}{c}{\textbf{Role}}
& \multicolumn{3}{c}{\textbf{Writing}}
& \multicolumn{2}{c}{\textbf{Social}}
& \multicolumn{2}{c}{\textbf{Instruct}}
& \multicolumn{3}{c}{\textbf{World}}
& \multicolumn{2}{c}{\textbf{Common}}
& \multicolumn{2}{c}{\textbf{Long}}
& \multicolumn{2}{c}{\textbf{Read}} \\

\cmidrule(lr){2-3}\cmidrule(lr){4-5}\cmidrule(lr){6-8}\cmidrule(lr){9-10}
\cmidrule(lr){11-12}\cmidrule(lr){13-15}\cmidrule(lr){16-17}\cmidrule(lr){18-19}
\cmidrule(lr){20-21}

%----------- dataset abbreviations (Reordered) -----
& Buzz & EQ3
& IG & RG
& CW & WB & JM
& SIQA & IQuiz
& IF & Collie
& MMLU & Hum & Soc
& HS & SC
& CoQA & CRS
& NQA & XSum
&  &  \\ \midrule

%===================== rows (Data Reordered) =====================%
Qwen3-8B
& 34.30 & 64.10                      % Emotion
& 58.71 & 74.40                      % Role
& 58.65 & 70.30 & 64.61             % Writing
& 75.44 & 57.81                      % Social
& 82.62 & 29.64 & 74.73 & 65.87 & \textbf{84.11} & 75.69 & 77.28 & 73.58 & 21.71 & 52.43 & 18.00 & 60.70 & 50.00 \\
 \midrule
 
Ultrafeedback
& 34.79 & 64.20                      % Emotion
& 61.07 & 76.72                      % Role
& 59.49 & 69.91 & 63.28             % Writing
& 74.92 & 55.00                      % Social
& 83.73 & 30.45 & \textbf{75.02} & \textbf{66.55} & 83.88 & 76.20 & 77.85 & 81.49 & 21.49 & 53.56 & 17.86 & 61.37 & 51.55 \\

WritingPrompts
& 33.71 & 63.25                      % Emotion
& 58.98 & 75.71                      % Role
& 61.47 & 70.17 & 63.35             % Writing
& 75.59 & 63.75                      % Social
& 83.92 & 30.93 & 74.83 & 66.08 & 83.59 & 76.12 & 77.72 & 81.53 & 21.51 & 53.01 & \textbf{18.08} & 61.67 & 52.30 \\

SynthQuestions
& 35.22 & 66.80                      % Emotion
& 63.92 & 77.91                      % Role
& 62.99 & 70.30 & 64.88             % Writing
& 74.72 & \textbf{65.00}             % Social
& 83.36 & 29.77 & 74.90 & 66.40 & 83.82 & 76.41 & 77.98 & 81.23 & 21.21 & 53.32 & 17.69 & 62.39 & 50.92 \\

Magpie\_Air
& 33.71 & 65.15                      % Emotion
& 61.88 & 77.09                      % Role
& 61.18 & 70.34 & 65.48             % Writing
& 75.38 & 57.50                      % Social
& 83.36 & 29.68 & 74.94 & 66.38 & 83.88 & \textbf{76.48} & 77.53 & \textbf{81.92} & 21.38 & 53.56 & 17.77 & 61.73 & 50.85 \\

Magpie\_Pro
& 34.35 & 65.85                      % Emotion
& 62.00 & 77.13                      % Role
& 62.96 & 70.14 & 63.80             % Writing
& 75.79 & 60.00                      % Social
& 83.18 & 30.33 & 74.93 & 66.52 & 83.95 & 76.34 & 77.91 & 81.39 & 21.09 & 53.76 & 17.91 & 61.97 & 51.12 \\

Magpie\_Mixed
& 33.61 & 65.50                      % Emotion
& 63.60 & 77.91                      % Role
& 61.48 & 70.48 & 64.26             % Writing
& 75.28 & 57.50                      % Social
& 84.05 & 31.43 & 74.99 & 66.32 & 84.08 & 76.17 & 77.83 & 81.86 & 21.66 & 53.88 & 18.02 & 62.00 & 52.78 \\

Self-Rewarding+M2
& 35.06 & 64.55                      % Emotion
& 58.37 & 73.77                      % Role
& 60.39 & 70.24 & 65.04             % Writing
& 75.13 & 57.50                      % Social
& 84.29 & 30.49 & 74.67 & 65.97 & 83.72 & 75.82 & 77.53 & 81.50 & 21.79 & \textbf{54.75} & 17.72 & 61.42 & 52.15 \\

Self-Rewarding+M3
& \textbf{36.60} & 65.65             % Emotion
& 61.95 & 76.55                      % Role
& 63.18 & 70.96 & 64.68             % Writing
& 74.46 & 60.00                      % Social
& \textbf{85.40} & 30.75 & 74.98 & 66.50 & 83.78 & 75.93 & 77.53 & 81.89 & 21.47 & 52.77 & 17.78 & 62.14 & 53.15 \\

FSC-wCL
& 33.99 & 64.05                      % Emotion
& 57.86 & 73.97                      % Role
& 58.61 & 69.99 & 63.35             % Writing
& \textbf{76.00} & 62.50             % Social
& 81.89 & 30.71 & 74.93 & 66.42 & 83.82 & 75.69 & 77.53 & 81.68 & 21.57 & 52.39 & 17.97 & 61.25 & 50.65 \\

FSC-woCL
& 33.54 & 64.45                      % Emotion
& 58.10 & 73.89                      % Role
& 58.30 & 70.10 & 62.90             % Writing
& 75.08 & 61.25                      % Social
& 83.92 & 30.68 & 74.97 & 66.42 & 83.91 & 75.65 & 77.47 & 81.53 & 21.63 & 52.12 & 17.80 & 61.19 & 51.95 \\

\midrule
\multicolumn{23}{l}{\textbf{BridgeAlign (Ours)}}\\
\rowcolor[gray]{0.96}\hspace{4pt} +RubricPO
& 34.05 & 65.80                      % Emotion
& 60.78 & 76.10                      % Role
& 62.76 & 70.60 & 64.07             % Writing
& 75.18 & 62.50                      % Social
& 82.99 & 31.44 & 74.95 & 66.42 & 83.85 & 75.83 & 78.04 & 80.99 & 21.65 & 52.70 & 17.90 & 61.93 & 51.80 \\

\rowcolor[gray]{0.96}\hspace{4pt} +BridgePO
& 35.75 & \textbf{67.15}             % Emotion
& \textbf{66.10} & \textbf{78.30}   % Role
& \textbf{64.00} & \textbf{71.05} & \textbf{65.65}    % Writing
& 75.20 & 64.92                      % Social
& 85.03 & 32.88 & 74.79 & 66.12 & 83.88 & \textbf{76.48} & \textbf{78.17} & 81.63 & 21.46 & 54.62 & 17.86 & \textbf{63.05} & \textbf{53.80} \\

\midrule
\multicolumn{23}{l}{\textbf{BridgeAlign (Ablation Study)}}\\
\hspace{4pt} +H-MPO
& 32.61 & 56.50                      % Emotion
& 57.00 & 72.68                      % Role
& 55.13 & 68.25 & 62.75             % Writing
& 74.87 & \textbf{65.00}             % Social
& 81.89 & 28.65 & 74.93 & 66.31 & 83.95 & 75.97 & 77.78 & 81.03 & \textbf{22.11} & 52.70 & 17.88 & 60.40 & 50.18 \\

\hspace{4pt} +BridgePO v1
& 35.75 & 64.75                      % Emotion
& 61.25 & 76.55                      % Role
& 61.17 & 70.49 & 63.81             % Writing
& 75.44 & 58.75                      % Social
& 83.92 & 30.18 & 74.88 & 66.25 & 83.69 & 75.91 & 77.53 & 80.55 & 21.55 & 53.15 & 17.92 & 61.67 & 51.48 \\

\hspace{4pt} +BridgePO v2
& 34.02 & 65.25                      % Emotion
& 59.76 & 74.98                      % Role
& 60.91 & 70.56 & 63.93             % Writing
& 74.67 & 55.00                      % Social
& 83.73 & 31.01 & 74.92 & 66.33 & 83.95 & 75.94 & 77.72 & 81.15 & 21.46 & 52.74 & 17.93 & 61.30 & 52.08 \\

\hspace{4pt} +BridgePO-Combo
& 34.13 & 65.65                      % Emotion
& 62.55 & 76.83                      % Role
& 59.60 & 70.29 & 64.57             % Writing
& 75.49 & 60.00                      % Social
& 82.44 & \textbf{32.91} & 74.88 & 66.16 & 83.88 & 75.92 & 77.91 & 81.12 & 21.76 & 53.08 & 17.67 & 61.84 & 52.62 \\

\bottomrule
\end{tabular}}
\caption{Full results on 17 mainstream benchmarks spanning nine core abilities (20 metrics in total), with AlpacaEval 2 listed separately. \textbf{Avg} is computed over the 20 benchmark metrics and excludes AlpacaEval 2. Higher values are better. \textit{Abbreviations:} Buzz = BuzzBench, EQ3 = EQ-Bench3, IG / RG = RoleBench\_IG\_En / RoleBench\_RG\_En, CW = CreativeWriting-v3, WB = WritingBench, JM = Judgemark-v2, SIQA = Social-IQA, IQuiz = IQuiz\_EQ, IF = IFEval, Hum / Soc = MMLU\_Humanities / MMLU\_Social\_Sciences, HS = HellaSwag, SC = StoryCloze, CRS = GovernmentReport\_CRS, NQA = NarrativeQA\_summarizeOnly, FSC-wCL / FSC-woCL = FollowSoftConstraint with / without curriculum learning, and BridgePO-Combo = BridgePO v1 + v2 + RubricPO. }
\label{tab:full_results}

\end{table*}

\begin{table*}[t]
\centering
\footnotesize
\setlength{\tabcolsep}{2pt}
\renewcommand{\arraystretch}{1.1}
\resizebox{1.0\textwidth}{!}{{
\begin{tabular}{lcccccccccccc}
\toprule
\multicolumn{1}{l}{\multirow{2}{*}{\textbf{Selected Method}}} &
\multicolumn{5}{c}{\textbf{Human Preference}} &
\multicolumn{6}{c}{\textbf{Knowledge based}} &
\multicolumn{1}{c}{\multirow{2}{*}{\textbf{Avg}}} \\
\cmidrule(lr){2-6}\cmidrule(lr){7-12}
& \textbf{Emotion}
& \textbf{Role}
& \textbf{Writing}
& \textbf{Social}
& \textbf{Avg HP}
& \textbf{Instruct}
& \textbf{World}
& \textbf{Common}
& \textbf{Long}
& \textbf{Read}
& \textbf{Avg KB} \\
\midrule
\rowcolor[gray]{0.97}BridgePO (800-step) & 51.73 & 71.99 & 66.31 & 66.53 & 64.38 & 56.91 & 75.01 & 76.74 & 51.98 & 35.13 & 60.60 & 62.30 \\
FollowSoftConstraint w/ CL & 49.02 & 65.92 & 63.98 & 69.25 & 62.26 & 56.30 & 75.06 & 76.61 & 51.63 & 35.18 & 60.42 & 61.25 \\
\midrule
\rowcolor[gray]{0.97}BridgePO (1200-step) & 52.36 & 72.19 & 66.87 & 66.54 & 64.75 & 57.30 & 75.07 & 76.86 & 52.01 & 35.59 & 60.79 & 62.57 \\
Self-Rewarding+M$_3$ & 51.13 & 69.25 & 66.27 & 67.23 & 63.78 & 58.08 & 75.09 & 76.73 & 51.68 & 35.28 & 60.80 & 62.14 \\
\bottomrule
\end{tabular}}}
\caption{{Near-matched token-budget controlled comparison. We compare BridgePO at 800 steps (442.3M tokens) against FollowSoftConstraint w/ CL (458.6M), and BridgePO at 1200 steps (663.5M) against Self-Rewarding+M$_3$ (648.1M).}}
\label{tab:token_matched_scaling}
\end{table*}

\begin{table*}[t]
\centering
\footnotesize
\setlength{\tabcolsep}{2pt}
\renewcommand{\arraystretch}{1.1}
\resizebox{1.0\textwidth}{!}{{
\begin{tabular}{lcccccccccccc}
\toprule
\multicolumn{1}{l}{\multirow{2}{*}{\textbf{Selected Method}}} &
\multicolumn{5}{c}{\textbf{Human Preference}} &
\multicolumn{6}{c}{\textbf{Knowledge based}} &
\multicolumn{1}{c}{\multirow{2}{*}{\textbf{Avg}}} \\
\cmidrule(lr){2-6}\cmidrule(lr){7-12}
& \textbf{Emotion}
& \textbf{Role}
& \textbf{Writing}
& \textbf{Social}
& \textbf{Avg HP}
& \textbf{Instruct}
& \textbf{World}
& \textbf{Common}
& \textbf{Long}
& \textbf{Read}
& \textbf{Avg KB} \\
\midrule
\rowcolor[gray]{0.97}RubricPO & 49.93 & 68.44 & 65.81 & 68.84 & 63.54 & 57.22 & 75.07 & 76.94 & 51.32 & 35.30 & 60.61 & 61.93 \\
w/o HSS Seeds+Generic Rubrics & 48.95 & 66.13 & 64.03 & 66.20 & 61.63 & 57.29 & 75.41 & 77.39 & 51.61 & 36.81 & 61.13 & 61.35 \\
\bottomrule
\end{tabular}}}
\caption{{HSS-specificity ablation under RubricPO. We compare RubricPO against an ablated variant that removes HSS-domain seeds and replaces HSS-specific scoring with generic rubrics.}}
\label{tab:hss_specificity_ablation}
\end{table*}

\section{{Controlled Ablations}}\label{sec:controlled_ablations_appendix}
{We conduct two additional controlled ablations to study HSS-specific supervision and token-budget effects. Let’s first clarify: BridgeAlign naturally produces longer and more diverse outputs because its instructions are reverse-generated from HSS seed documents while preserving domain, genre, length, and persona constraints. We regard these properties as intended features of the pipeline rather than simple confounders, and explicitly examine them below.}

\paragraph{Token-Budget Controlled Comparison}
Under the same 210k-sample setting, BridgeAlign +BridgePO contains 910.6M training tokens, whereas Self-Rewarding+M$_3$ and FollowSoftConstraint w/ CL contain 648.1M and 458.6M tokens, respectively, indicating that matching sample counts alone is insufficient. To control for token-budget differences, in Table~\ref{tab:token_matched_scaling}, we compare two intermediate BridgePO checkpoints with the same-setup baselines at nearby token budgets: the 800-step checkpoint (442.3M tokens) against FollowSoftConstraint w/ CL, and the 1200-step checkpoint (663.5M tokens) against Self-Rewarding+M$_3$.
BridgePO achieves stronger human-preference performance in both comparisons, with Avg HP of 64.38 vs. 62.26 and 64.75 vs. 63.78, respectively. Meanwhile, knowledge-side performance remains nearly unchanged, with Avg KB of 60.60 vs. 60.42 and 60.79 vs. 60.80. These results show that the gains are not driven by a larger token budget.

\paragraph{{HSS-Specificity Ablation}}
To separate HSS-specific effects from generic text-quality effects, we remove HSS-domain seeds and replace HSS-specific scoring with generic rubrics, constructing data constrained only by general text-quality criteria. Since BridgePO relies on controlled degradation along the human-touch dimension, this ablation cannot be directly applied to BridgePO and is therefore conducted in Table~\ref{tab:hss_specificity_ablation} under the RubricPO setting.
{Removing HSS seeds and HSS-oriented scoring reduces Avg HP from 63.54 to 61.63 and overall Avg from 61.93 to 61.35. The main drops occur on HSS-oriented evaluations: Emotion, Role, Writing, and Social decrease by 0.98, 2.31, 1.78, and 2.64, respectively. In contrast, knowledge-side metrics improve slightly, raising Avg KB from 60.61 to 61.13. This pattern suggests that HSS seed construction and HSS-specific rubrics provide targeted gains for HSS-related capabilities rather than merely improving generic text quality.}

\begin{figure*}[t]
\centering
\includegraphics[width=\textwidth]{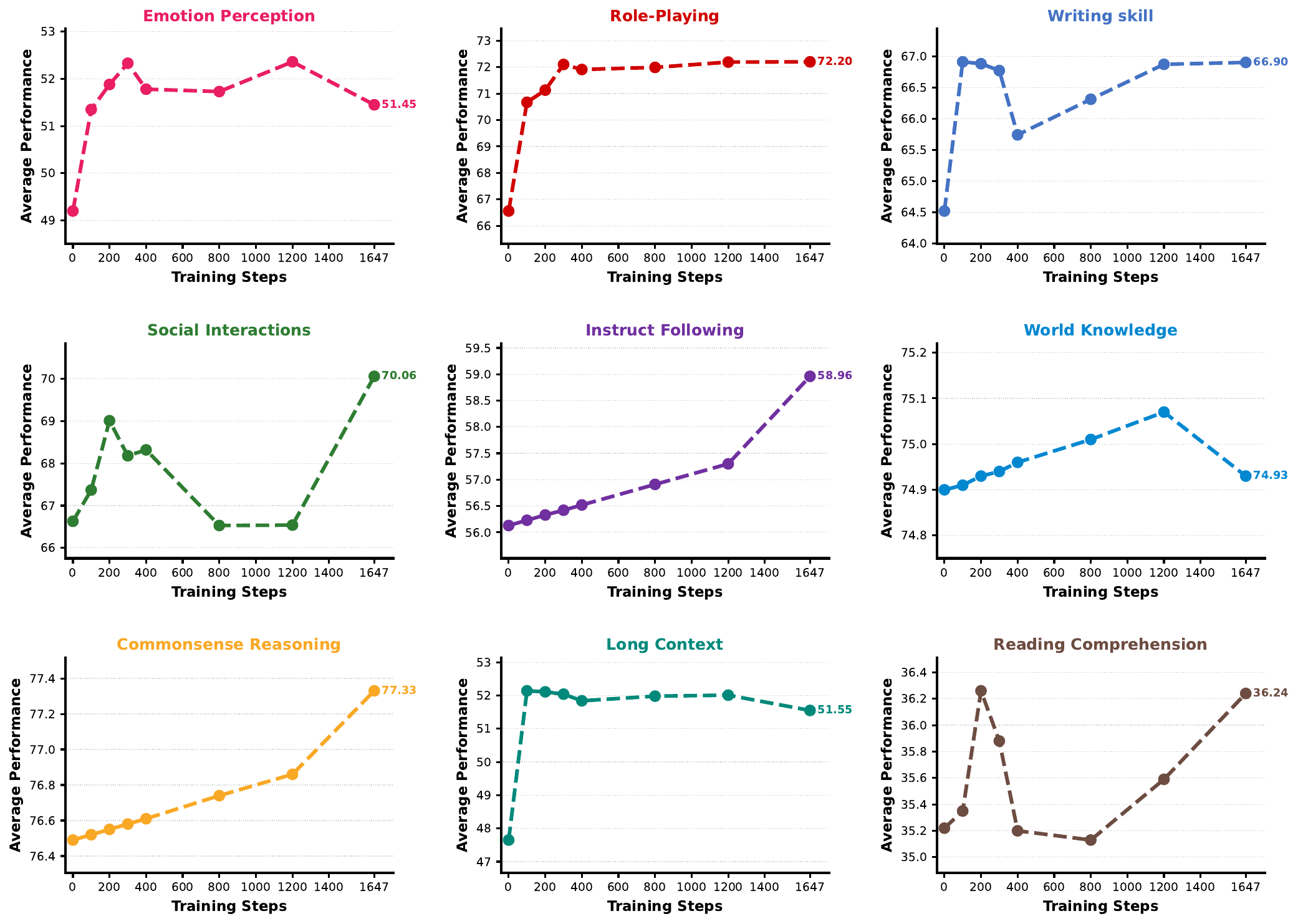}
\caption{Per-capability data scaling curves for BridgePO over one epoch. Emotion perception, writing skill, and long context peak early and then decline slightly, suggesting mild overfitting; role-playing shows only slight fluctuations after the initial gain; reading comprehension and social interaction rise early, dip in the middle, and rebound toward the end; instruction following and commonsense reasoning keep improving, rising steadily toward the end; and world knowledge stays largely flat throughout.}
\label{fig:scaling_effect_9panel}
\end{figure*}

\section{{Theoretical Interpretation}}\label{sec:theory}

This section draws on existing theoretical analyses of LLM data synthesis and preference optimization~\cite{gan2025towards,rafailov2024direct} to interpret the empirical patterns observed in BridgeAlign experiments.

\paragraph{Reverse-Bottleneck in Data Synthesis.} 
Following the \textbf{Reverse-Bottleneck} perspective in \cite{gan2025towards}, the data synthesis process of BridgeAlign can be interpreted as introducing an information gain $\Delta I$ through the synthesis model $M$:

\begin{equation}
\Delta I = H(M(p)) - I(h(e_p), M(p)). \label{eq:info_gain}
\end{equation}

Here, $H(M(p))$ is the entropy of the generated responses, $I(h(e_p), M(p))$ is the mutual information between the responses and the prompt $p$. Thus, $\Delta I$ measures the incremental information injected by $M$ from $p$ to the synthetic data $S_{\text{gen}}$.
According to Theorem 4.7 in \cite{gan2025towards}, the generalization error of the model $\pi$ trained on synthetic data is bounded by: 

\begin{equation}
\resizebox{0.5\textwidth}{!}{
$
\begin{aligned}
\mathbb{E}\!\left[\mathrm{Err}(\pi^{S_{\text{gen}}})\right] \;\le\;& C \Big( D_{\mathrm{TV}}(\mathcal{D}, \mathcal{D}_M) + D_{\mathrm{TV}}(\mathcal{D}_M, \mathcal{D}_{\text{gen}}) \Big) 
\\
&+ \exp\!\left( -\frac{L}{2} \log \frac{1}{\eta} \right) \sqrt{ \frac{ 2\sigma^2 \big[ -\Delta I + B_{\text{syn}} + H(e_M) + \delta_{\epsilon,p} \big] }{n} }.
\end{aligned}
$}
\end{equation}
The generalization error is constrained by the interaction between the $\Delta I$ and the \textbf{compression bottleneck $B_{\text{syn}}$}. 
While $B_{\text{syn}}$ represents the inherent limit of information compression during the learning process, the term $-\Delta I$ implies that maximizing information gain through high-quality synthesis directly reduces the generalization error upper bound. 
Furthermore, the \textbf{Generalized Generalization Model Inference (GGMI)} analysis (Theorem 4.10 in \cite{gan2025towards}) quantifies the generalization gain benefit from synthetic data:
\begin{equation}
\begin{aligned}
\mathrm{GGMI} \;\le\;& \Delta I - (\alpha + 1)\, H(S_{\text{anchor}} \mid W) 
\\ 
&+ 2 \Delta H + H(S_{\text{gen}} \mid W) + \epsilon_{W,p}.
\end{aligned}
\end{equation}
In this interpretation, $\Delta I$ captures the diversity introduced by the synthesis model, while $\Delta H = H(S_{\text{anchor}}) - H(S_{\text{gen}})$ captures the reduction in uncertainty brought by filtering and refinement. From this perspective, BridgeAlign may benefit from combining broader coverage with cleaner training signals.

\paragraph{Implicit Reward Alignment via DPO.} 
Our preference optimization stage uses the standard DPO formulation, which can be viewed as a KL-constrained reward maximization objective~\cite{rafailov2024direct}:
\begin{equation}
\begin{aligned}
\max_{\pi_\theta}\; & \mathbb{E}_{x \sim \mathcal{D},\, y \sim \pi_\theta(\cdot \mid x)} \big[ r_\phi(x, y) \big] \\ &- \beta\, \mathbb{D}_{\mathrm{KL}} \!\left( \pi_\theta(y \mid x) \,\Vert\, \pi_{\text{ref}}(y \mid x) \right).
\end{aligned}
\end{equation}
By applying the Bradley-Terry model and reparameterization, the ground-truth reward $r$ and the optimal policy $\pi_r$ satisfy: 
\begin{equation}r(x, y) = \beta \log \frac{\pi_r(y|x)}{\pi_{\text{ref}}(y|x)} + \beta \log Z(x)\end{equation}
This allows us to fit an implicit reward by minimizing the DPO loss on our synthetic triplets $\langle x, y_w, y_l \rangle$ like Eq.\ref{eq:dpo_objective}.
Under this view, preference optimization improves the policy by enlarging the implicit reward gap between preferred and dispreferred responses. BridgeAlign, especially BridgePO, promotes fine-grained quality discrimination by constructing more challenging near-boundary comparisons.

\section{Case Study}
Tables~\ref{tab:case_study} and~\ref{tab:case_study_part2} present an arts-domain case from the BridgeAlign synthesis pipeline, including the seed document, inverted instruction, inverted document, and transitional document. Its metadata include the genre \textit{Essay}, the persona ``a discerning classical music critic with a focus on performance analysis,'' and context lengths of 424, 252, 593, and 466 tokens for the four components, respectively.

\section{Limitations}
BridgeAlign constructs preference data from web corpora through a three-phase pipeline. Despite heuristic filtering, LLM-signal filtering, and LLM-based rewriting, the web data may still contain crawler noise and low-quality expressions. Future work should improve seed data quality by adopting stronger cleaning techniques, using cleaner web datasets such as FineWeb, or purchasing high-quality human-written corpora.

Data-scaling studies further show that the model largely converges after about 51k preference samples, with sharply diminishing marginal gains. This early saturation may reflect potential redundancy, but it remains unclear whether it is specific to Qwen3-8B, persists across model scales, or indicates the need for diversity beyond semantic and domain coverage. Future work will reduce potential redundancy through stricter semantic deduplication, such as embedding-based similarity filtering or clustering-based sampling.

Finally, although Appendix Table~\ref{tab:domain_composition} shows BridgeAlign’s coverage of 14 domains, its domain distribution remains imbalanced. This long-tail pattern likely reflects differences in web-text availability across HSS fields rather than intentional sampling bias, but it complicates per-domain training and evaluation, especially for low-resource domains. We therefore report domain statistics for transparency and leave systematic per-domain evaluation to future work.

\begin{table*}[t]
\small
\setlength{\tabcolsep}{2pt}
\begin{tabularx}{\textwidth}{lX}
\toprule
\textbf{Text Type} & \textbf{Text}  \\
\midrule
Seed \\Document & \detokenize{This was a typically imaginative programme from Nicolas Hodges, and one that cast a spellbinding sound world over proceedings. If the Lachenmann's sonata' had been programmed for its undoubted hypnotic strengths, then Hodges generated equal hypnotism in his playing of Schumann - particularly the Fantasie op.17 which was given a masterful performance.\n\nLachenmann's Serynade is a provocative work which marries violent clusters and distilled nervousness with apparent equilibrium. However, there is little that is eclectic about its post-modernist harmony and it by no means exhausts the possibilities of either the instrument or the pianist's technique. Indeed, its half-hour span can grate slightly - the work having an almost obsessive desire to hypnotise the listener with endlessly repeated chords. The sustained use of different keys and the various pedal techniques encourage the mood of hypnotism - and the piece almost works in removing the listener from the conscious world. Hodges' playing was faultless and he brought transcendent colour to his tone.\n\nWhilst Hodges' displayed no lack of involvement in the Lachenmann he was more persuasive in Schumann. The Arabeske op.18 is dark in colour, the pianist's hands predominantly playing only on the lower half of the keyboard. There were some beautiful dynamic touches - for example, Hodges' fingers caressing, like paws, the notes where Schumann had indicated p, and exploding with a tiger's power to the bass clusters. In the longer Fantasie he worked wonders with Schumann's imagist writing. There was much beauty of phrasing and panache to the virtuoso setting of the second movement, which momentarily broke the trance he had carefully constructed throughout its development. The poetry and lyricism given to the work were instinctive, although the playing appeared unusually direct for Schumann. If the mood gestured unmistakably towards restrained passion this seemed right for the moment. If there was one fault it is his tendency to over-pedal and this was occasionally intrusive in both Schumann works. A fine recital.}\\
\midrule
Inverted \\Instruction & \detokenize{You are a discerning classical music critic with a focus on performance analysis. Your stance prioritizes technical precision and emotional authenticity, valuing works that challenge conventional structures while maintaining lyrical depth. Your mindset is analytical yet reflective, balancing admiration for virtuosity with a critical eye for overused techniques. The tone remains measured and insightful, emphasizing the interplay between composer intent and performer interpretation.\nWrite an Arts essay in the Genre of Essay that analyzes a pianist's interpretation of modern classical compositions, focusing on the interplay between technical execution and emotional expression. Summarize the central thesis that a performer's mastery lies in balancing hypnotic repetition with dynamic nuance, using specific examples from a Lachenmann sonata and Schumann works. Include key points such as the use of clusters, pedal techniques, and contrasting dynamics, while retaining exact references to the pieces (e.g., 'Fantasie op.17', 'Arabeske op.18') and the performer's stylistic choices. Structure the essay thematically, comparing the works' structural demands and the pianist's approach to each. The final text must be approximately 424 tokens (around 2088 characters), matching the length of the original document.} \\
\bottomrule
\end{tabularx}
\caption{
A case from the BridgeAlign synthesis pipeline (Part 1 of 2): the seed document and the inverted instruction.
}\label{tab:case_study}
\end{table*}

\section{Prompt Template}\label{sec:prompt_template}
This section outlines the prompt designs adopted throughout the pipeline.
Figure~\ref{fig:hss_domain_cls_prompt} presents the HSS domain classification prompt, which assigns each seed document to its corresponding HSS domain.
Figures~\ref{fig:hss_quality_filter_part1_prompt}–\ref{fig:hss_quality_filter_part3_prompt} illustrate the HSS quality filtering prompts. Here, using expert-crafted rubrics to retain only clean-filtered seeds.
The text refinement and refinement evaluation templates in Figures~\ref{fig:hss_text_refinement_prompt} and \ref{fig:hss_refinement_evaluation_prompt} are used for text refining and for evaluating the refined documents, respectively.
Figure~\ref{fig:hss_instruction_inversion_prompt} shows the instruction inversion prompt that reverse-generates an instruction from a seed document, whereas Figure~\ref{fig:hss_qa_consistency_check_prompt} depicts the Q\&A consistency check prompt that verifies the alignment of the inverted instruction between its origin (i.e, seed document).
Figure \ref{fig:hss_controlled_quality_degradation_prompt} shows the prompt template for controlled quality degradation of the higher-scoring document (either seed or inverted) alongside a human-touch dimension to produce a transitional document.

\begin{table*}[t]
\small
\setlength{\tabcolsep}{2pt}
\begin{tabularx}{\textwidth}{lX}
\toprule
\textbf{Text Type} & \textbf{Text}  \\
\midrule
Inverted \\Document & \detokenize{In the realm of modern classical performance, the pianist\u2019s mastery lies in navigating the tension between hypnotic repetition and dynamic nuance, a balance exemplified in Nicolas Hodges\u2019 interpretation of Lachenmann\u2019s *Serynade* and Schumann\u2019s *Fantasie op. 17*, *Arabeske op. 18*. These works demand not mere technical facility but a philosophical engagement with structure and emotion, where the performer becomes both architect and alchemist.  \n\nLachenmann\u2019s *Serynade*, with its postmodernist dissonance, hinges on clusters and sustained pedal techniques to evoke a trance-like state. Hodges\u2019 execution of the sonata\u2019s repetitive chord progressions was meticulous, his hands weaving through violent, staccato bursts that juxtaposed with the piece\u2019s eerie equilibrium. The relentless oscillation between minor and major keys, amplified by layered pedaling, created a sonic fog that both entranced and unsettled. Yet, the work\u2019s endurance\u2014its half-hour sprawl\u2014rested on Hodges\u2019 ability to infuse each repetition with subtle inflection, preventing monotony. His use of the sustaining pedal, for instance, did not merely blur sonorities but sculpted a spatial depth, as if the piano itself became a vessel for psychological tension.  \n\nContrast this with Schumann\u2019s *Fantasie op. 17*, where Hodges\u2019 approach revealed a different kind of duality. The piece\u2019s fluid, improvisatory structure required a delicate interplay between lyrical phrasing and abrupt dynamic shifts. In the *Arabeske op. 18*, his hands lingered in the lower register, evoking a somber, almost feral energy. The pianist\u2019s treatment of Schumann\u2019s *p* markings\u2014fingers \u201ccaressing\u201d notes with precision\u2014contrasted starkly with the explosive bass clusters, a chiaroscuro of restraint and power. Here, the hypnotic element emerged not through repetition but through the interplay of light and shadow, as Hodges\u2019 phrasing danced between fragility and intensity.  \n\nStructurally, both works challenged Hodges to reconcile opposing forces: Lachenmann\u2019s obsessive formalism versus Schumann\u2019s romantic spontaneity. Yet his stylistic choices\u2014whether the calculated detachment in *Serynade* or the fervent directness in Schumann\u2014revealed a performer attuned to each composer\u2019s ethos. Even his occasional over-pedaling, which occasionally muddied the Schumann works, underscored the risks of prioritizing atmosphere over clarity, a trade-off that mirrored the composers\u2019 own tensions between innovation and tradition.  \n\nUltimately, Hodges\u2019 performance affirmed that technical precision and emotional authenticity are not adversaries but collaborators. In balancing the hypnotic and the dynamic, he transformed these works into meditations on time, tension, and the fragile beauty of impermanence.} \\
\midrule
Transitional \\Document & \detokenize{In analyzing pianist Hodges\u2019s interpretation of Lachenmann\u2019s *Serynade* and Schumann\u2019s *Fantasie op. 17* and *Arabeske op. 18*, the central thesis emerges: mastery lies in balancing hypnotic repetition with dynamic nuance, where technical precision and emotional authenticity coalesce. Lachenmann\u2019s sonata, a labyrinth of dissonance and fragmentation, demands a performer who navigates clusters and extended techniques without sacrificing lyrical intent. Hodges employs a meticulous, almost surgical approach to the work\u2019s repetitive motifs, such as the relentless ostinato in the first movement, yet infuses them with subtle gradations of touch, transforming mechanical repetition into a meditative, almost ritualistic experience. The use of the sustain pedal here is transformative\u2014blurring harmonic boundaries to evoke a spectral, dreamlike atmosphere, a technique that aligns with Lachenmann\u2019s intent to destabilize traditional tonality.  \n\nContrastingly, Schumann\u2019s *Fantasie op. 17* and *Arabeske op. 18* require a pianist to negotiate lyrical fluidity and structural tension. Hodges\u2019s reading of the *Fantasie* emphasizes dynamic contrasts, particularly in the interplay between the stormy, chromatic passages and the introspective, bell-like chords of the second movement. The performer\u2019s use of rubato is restrained yet expressive, allowing phrases to breathe without sacrificing rhythmic cohesion. In the *Arabeske*, Hodges balances the piece\u2019s delicate ornamentation with a taut, forward-driving pulse, ensuring that the work\u2019s elegance never succumbs to passivity.  \n\nThe interplay between Lachenmann\u2019s avant-garde structures and Schumann\u2019s Romantic lyricism reveals Hodges\u2019s versatility. Where Lachenmann\u2019s clusters demand a focus on timbral exploration, Schumann\u2019s harmonies require a sensitivity to textural layering. Hodges\u2019s ability to oscillate between these extremes\u2014maintaining technical rigor while infusing each note with emotional resonance\u2014defines their interpretation. This duality underscores the performer\u2019s mastery: not as a mere executor of notation, but as a mediator between the composer\u2019s vision and the listener\u2019s experience.} \\
\bottomrule
\end{tabularx}
\caption{
A case from the BridgeAlign synthesis pipeline (Part 2 of 2): the inverted document and the transitional document.
}\label{tab:case_study_part2}
\end{table*}

\begin{figure*}[t]
\centering
\begin{promptbox}{Prompt Template for {HSS Domain Classification}}
You are an expert in text domain classification.
Your task is to classify the given text into \textbf{one} of the fourteen predefined domain types, or determine if it falls outside these domains. Accuracy in identifying non-relevant texts is as important as correctly classifying relevant texts.\\

\texttt{[Text]}: \textcolor{red}{\textbf{\{text\}}}\\

Predefined Domain Types: [Philosophy, Economics, Law, Politics, Sociology, Health and Nursing, Geography, Education, Sports, Literature, History, Management, Arts, Psychology]\\

\texttt{[Instruction]} \\
\begin{enumerate}[topsep=2pt, itemsep=4pt, parsep=0pt, leftmargin=14pt]
\item Carefully analyze the main subject matter, methodology, and terminology of the text.

\item Determine if the text's **primary focus** clearly and substantially aligns with \textbf{one} of the domains listed above.

\item \textbf{If a strong match is found}: Select the \textbf{single best-fitting domain} from the list.

\item \textbf{Crucially: If the text does not clearly and primarily belong to any of the 14 listed domains} (e.g., it is about natural sciences, technology, engineering, mathematics, a general news report without relevant domain focus, creative writing without clear thematic analysis, a personal blog, technical documentation, etc.), you \textbf{MUST} output ``None'' in the ``Domain Types'' field. \textbf{Do NOT force a classification} into the listed domains if the fit is poor or ambiguous.

\item Rate your confidence in the assigned domain (or the ``None'' classification) on a scale of 1 (lowest confidence, very uncertain or weak fit) to 5 (highest confidence, very clear and strong fit/non-fit). A low confidence score (1-2) for a chosen domain might indicate it's a borderline case.

\item Output the result \textbf{strictly} in the specified JSON format. Generate only the JSON object, with no additional text before or after it.\\
\end{enumerate}

\texttt{[Output Format]} \\
\{\\
  ``Domain Types'': ``<Selected Domain or None>'', \\
  ``Confidence'': <1-5>\\
\}
\end{promptbox}
\caption{Prompt template for HSS domain classification.}
\label{fig:hss_domain_cls_prompt}
\end{figure*}

\begin{figure*}[t]
\centering
\begin{promptbox}{Prompt Template for {HSS Quality Filtering} (Part I)}
You are a professional document quality reviewer; if the text warrants it, do not hesitate to use the full range of scores, including the lowest and highest scores.
Evaluate the document based on the following criteria, each with a maximum score of 5 points. Please provide a brief explanation for each criterion and output the final result in JSON format directly, where the score represents the score for each item, and the reason provides a simple explanation. Besides the genre, which is given separately.

\begin{itemize}[topsep=0pt, itemsep=4pt, parsep=0pt, leftmargin=12pt]
  \item \textbf{Grammar, Punctuation, and Spelling}: Assess whether there are noticeable grammatical errors, improper punctuation use, or spelling mistakes in the article. Also, check if there's any garbled text or illogical multilingual mixing.
  \begin{itemize}[nosep, leftmargin=10pt]
    \item 5: No errors at all.
    \item 4: Minor errors only.
    \item 3: Some noticeable errors, but understandable.
    \item 2: Many errors affecting the reading experience.
    \item 1: Errors are pervasive and severely impact text quality, or the text includes extensive meaningless repetition, gibberish, or a mix of illogical language, rendering it largely unreadable.
  \end{itemize}
  
  \item \textbf{Logical Coherence and Fluency}: Examine whether the text has a good logical structure and if the transitions between paragraphs are smooth.
  \begin{itemize}[nosep, leftmargin=10pt]
    \item 5: Very clear logic and fluent.
    \item 4: Mostly logical, some abrupt parts.
    \item 3: Overall acceptable but with a few noticeable gaps.
    \item 2: Logic is relatively disorganized, making it difficult to follow the author's line of thought, or there is noticeable redundancy or unnecessary repetition that interferes with reading fluency.
    \item 1: Completely lacks organization, or due to extreme, meaningless repetition, internal contradictions, or a complete structural breakdown, the text's logic completely fails, rendering it incomprehensible.
  \end{itemize}
  
  \item \textbf{Content Accuracy}: Ensure information is accurate and tightly centered around the topic.
  \begin{itemize}[nosep, leftmargin=10pt]
    \item 5: All information is very accurate and closely centered on the main idea.
    \item 4: Mostly correct with only a few details needing correction.
    \item 3: Some content deviates from the topic or contains small errors.
    \item 2: Many important factual errors.
    \item 1: Grossly inaccurate.
  \end{itemize}
  
  \item \textbf{Domain Relevance}: Assess whether the text's content, terminology, concepts, and focus are closely related to the specific academic field it claims or implies (e.g., history, sociology, literature, philosophy, etc.). Predefined Domain Types: [Philosophy, Economics, Law, Politics, Sociology, Health and Nursing, Geography, Education, Sports, Literature, History, Management, Arts, Psychology] The domain of the document is \textcolor{red}{\textbf{\{Domain\}}}.
  \begin{itemize}[nosep, leftmargin=10pt]
    \item 5: Highly relevant. The text fully focuses on the core issues, concepts, and terminology of the specified field, clearly belonging to that field.
    \item 4: Good relevance. The main focus is appropriate, but it may contain some marginal content or slight deviations.
    \item 3: Moderately relevant. The text touches on the field but may include more unrelated content, or the focus is too broad, lacking specific connections to the field.
    \item 2: Weak relevance. The connection to the field is weak or tenuous, with most content seeming unrelated to the topic.
    \item 1: Almost non-relevant. The text content has little or no connection to the specified field.
  \end{itemize}

  \item \textbf{Literary Diversity}: Evaluate the diversity and innovation in rhetorical devices (metaphors, personification, exaggeration, etc.), expression methods (narrative, exposition, argumentation, etc.), article structure (transition, flashback, etc.), narrative perspective, and person.
  \begin{itemize}[nosep, leftmargin=10pt]
    \item 5: The work shows high diversity and innovation in various aspects, making the text rich and vivid.
    \item 4: Good diversity in most areas with some slight monotony or traditionalism.
    \item 3: Some diversity but overall rather plain, lacking enough innovation.
    \item 2: Poor diversity, and both rhetorical devices and expression methods are monotonous, with traditional structure and perspective.
    \item 1: Almost no diversity, rhetorical devices and expression methods are very monotonous, and the structure and perspective are dull.
  \end{itemize}
\end{itemize}

\textit{\textbf{(Continue in the next page)}}

\end{promptbox}
\caption{Prompt Template for HSS Quality Filtering (Part I).}
\label{fig:hss_quality_filter_part1_prompt}
\end{figure*}

\begin{figure*}[t]
\centering
\begin{promptbox}{Prompt Template for {HSS Quality Filtering (Part II)}}

\begin{itemize}[topsep=0pt, itemsep=4pt, parsep=0pt, leftmargin=12pt]
  \item \textbf{Vocabulary Richness}: Check if the article uses varied vocabulary instead of repeatedly using the same words.
  \begin{itemize}[nosep, leftmargin=10pt]
    \item 5: Extremely diverse vocabulary.
    \item 4: Mostly rich vocabulary.
    \item 3: Basically sufficient but somewhat monotonous.
    \item 2: Quite limited.
    \item 1: Extremely limited vocabulary; heavy reliance on a very restricted set of words, or features extreme, meaningless repetition of words/phrases, completely sacrificing expressive richness.
  \end{itemize}
  
  \item \textbf{Knowledge Depth and Breadth}: Measure the degree of expertise and detail in the background information provided.
  \begin{itemize}[nosep, leftmargin=10pt]
    \item 5: Provides a wealth of in-depth and detailed related knowledge.
    \item 4: Quite comprehensive information but lacks depth.
    \item 3: Basic introduction is relatively sufficient.
    \item 2: Overly brief.
    \item 1: Almost no additional information.
  \end{itemize}
  
  \item \textbf{Theme Exploration Depth}: Evaluate whether the work deeply explores its core issue rather than skimming the surface.
  \begin{itemize}[nosep, leftmargin=10pt]
    \item 5: Profound insights and thought-provoking.
    \item 4: Some depth but could still go further.
    \item 3: Remains at a superficial discussion.
    \item 2: Very superficial treatment.
    \item 1: Did not touch the essence at all.
  \end{itemize}

    \item \textbf{Humanities Creativity}: Different from the creativity of scientific and engineering works, the creativity of humanities works is about observing the world from a unique perspective, designing plots and themes innovatively, and using creative expression techniques to convey thoughts and emotions, thereby inspiring readers or audiences to think and feel, bringing new value to human culture and the spiritual world.
    \begin{itemize}[nosep, leftmargin=10pt]
      \item 5: Highly creative. The work exhibits unprecedented novelty in view, plot, and achieves a high level of innovation in artistic style and expression, having a profound impact on the cultural field;
      \item 4: Quite creative. Exhibits a high degree of originality and uniqueness in content or form, effectively capturing readers' attention and provoking deeper thoughts;
      \item 3: Basic creativity. Can propose interesting view or relatively novel forms from more common perspectives but has not reached the level of being refreshingly new;
      \item 2: Some attempts but not very successful. Though there have been some innovative attempts, the overall impression remains quite ordinary, leaving no deep impression;
      \item 1: Almost no creativity. Content is outdated, lacks novelty, and the expression technique is mediocre, unable to arouse readers' interest or thought.
    \end{itemize}
  
    \item \textbf{Genre Recognition}: Identify the genre of the document, such as poetry, fiction, essay, advertisement, news report, public document, official document, literature, academic paper, report, etc.

    \textbf{Note}: The scores for ``Tone and Expression'', ``Emotional Expression'', and ``Genre Focus'' must relate to the genre.
    
    \item \textbf{Tone and Expression}:
    \begin{itemize}[nosep, leftmargin=10pt]
      \item 5: Perfectly matches the required style for the genre.
      For advertisements, essays, novels, plays, and screenplays, speeches, etc.: Uses appropriate tone words, colloquial or humorous expressions.
      For news reports, official documents, literature, academic papers, reports, etc.: Uses precise wording, maintaining written expression as much as possible rational and objective.
      \item 4: Mostly meets the requirements with only slight deviations from the expected style.
      \item 3: Expression method basically meets genre needs but with some inconsistencies.
      \item 2: Significantly deviates from the expected style but still partially acceptable.
      \item 1: Almost entirely does not meet the expression style required by the genre.
    \end{itemize}
\end{itemize}
\textit{\textbf{(Continue in the next page)}}
\end{promptbox}
\caption{Prompt Template for HSS Quality Filtering (Part II).}
\label{fig:hss_quality_filter_part2_prompt}
\end{figure*}

\begin{figure*}[t]
\centering
\begin{promptbox}{Prompt Template for {HSS Quality Filtering (Part III)}}
\begin{itemize}[topsep=0pt, itemsep=4pt, parsep=0pt, leftmargin=12pt]

    \item \textbf{Emotional Expression}:
    \begin{itemize}[nosep, leftmargin=10pt]
      \item 5: Emotionally rich and appropriate, perfectly matching the emotional depth required by the genre.
      For advertisements, essays, novels, plays, and screenplays, speeches, etc.: Emotion is abundant and appropriate.
      For news reports, official documents, literature, academic papers, reports, etc.: Uses precise wording to minimize personal emotional expression.
      \item 4: Emotionally appropriate with minor shortcomings or excesses.
      \item 3: Medium level, neither particularly outstanding nor notably flawed.
      \item 2: Weak emotional expression, or overly emotional for certain genres.
      \item 1: Extremely lacking in appropriate emotional color, or severely improper emotional expression.
    \end{itemize}
    
    \item \textbf{Genre Focus}:
    \begin{itemize}[nosep, leftmargin=10pt]
      \item 5: The document is highly consistent with the genre's focus in structure, language, and overall presentation, fully meeting the requirements of its genre.
      Announcements, notices, contracts need to emphasize rigor and professionalism, especially in standardized format, ensuring accuracy and ease of understanding.
      Reports, plans, and summaries require clarity, detailed content, and some format norms to reflect their seriousness.
      Regulations demand clarity and explicit statement for ease of compliance.
      Speeches pursue language appeal and expressiveness to resonate with the audience, and language expression needs to be colloquial with a certain diversity (writing techniques, rhetorical diversity, structure diversity, vocabulary diversity, etc.), but does not require the same level of diversity as pure literary creation.
      Literary creations such as novels, poetry, and essays focus more on emotional expression and aesthetic beauty, attracting readers through rich plot settings and graceful language style and requiring ample diversity (writing techniques, rhetorical diversity, structure diversity, etc.).
      Biographies and historical novels need to ensure both story appeal and factual accuracy with necessary diversity (writing techniques, rhetorical diversity, structure diversity, etc.).
      Sci-fi and fantasy rely on novel and unique world settings to spark readers' imagination, needing to abide by certain scientific laws and requiring necessary diversity (writing techniques, rhetorical diversity, structure diversity, etc.).
      Travelogues and essays record personal experiences or insights, reflecting personal color, and require necessary diversity (writing techniques, rhetorical diversity, structure diversity, etc.).
      \item 4: Generally good performance but slightly lacking in some details.
      \item 3: Basically meets requirements but with clear room for improvement.
      \item 2: Fails to meet expected standards in multiple key areas.
      \item 1: Severely violates the basic rules or conventions of the genre.
    \end{itemize}
\end{itemize}

The domain of the document is \textcolor{red}{\textbf{\{Domain\}}}.\\
\texttt{[Input Text]}: \textcolor{red}{\textbf{\{text\}}}\\

\textbf{Example Output:}
\begin{verbatim}
{
    "Grammar, Punctuation, and Spelling": {"score": <1-5>, "reason": "<detailed explanation>"},
    "Logical Coherence and Fluency": {"score": <1-5>, "reason": "<detailed explanation>"},
    "Content Accuracy": {"score": <1-5>, "reason": "<detailed explanation>"},
    "Domain Relevance": {"score": <1-5>, "reason": "<detailed explanation>"},
    "Literary Diversity": {"score": <1-5>, "reason": "<detailed explanation>"},
    "Vocabulary Richness": {"score": <1-5>, "reason": "<detailed explanation>"},
    "Knowledge Depth and Breadth": {"score": <1-5>, "reason": "<detailed explanation>"},
    "Theme Exploration Depth": {"score": <1-5>, "reason": "<detailed explanation>"},
    "Humanities Creativity": {"score": <1-5>, "reason": "<detailed explanation>"},
    "Genre": "",
    "Tone and Expression": {"score": <1-5>, "reason": "<detailed explanation>"},
    "Emotional Expression": {"score": <1-5>, "reason": "<detailed explanation>"},
    "Genre Focus": {"score": <1-5>, "reason": "<detailed explanation>"}
}
\end{verbatim}

\end{promptbox}
\caption{Prompt Template for HSS Quality Filtering (Part III).}
\label{fig:hss_quality_filter_part3_prompt}
\end{figure*}

\begin{figure*}[t]
\centering
\begin{promptbox}{Prompt Template for {Text Refinement}}
You are an expert in text refinement and content cleansing, specializing in preparing ``answer'' texts for high-quality datasets.\\

Your core task is to: thoroughly refine and clean the raw text provided below, to maximally eliminate all ``crawler traces'' and any redundant information unrelated to the article's core content, while \textbf{absolutely ensuring 100\% completeness and accuracy of the original text's core content, structure, meaning, author's tone, and ``human touch,''} enabling it to serve directly as a reference answer for Training.\\

\texttt{[Specific Operational Requirements]}
\begin{enumerate}[topsep=2pt, itemsep=4pt, parsep=0pt, leftmargin=14pt]
  \item \textbf{Thoroughly Clear Useless Information (Crawler Traces \& Redundancy):}
        \begin{itemize}[nosep, leftmargin=10pt]
            \item \textbf{Must remove, including but not limited to:}
                  \begin{itemize}[leftmargin=*, labelsep=0.5em, itemsep=2pt]
                      \item Webpage navigation bars (headers, footers, sidebars, menus, breadcrumb navigation)
                      \item Book tables of contents (unless the content itself is part of a TOC structure and must be retained)
                      \item Publisher information, copyright notices, disclaimers
                      \item Advertisements, promotional material, related article recommendations, social sharing buttons/prompts
                      \item Subscription prompts, user comment sections, donation requests
                      \item Excessive blank lines, residual broken HTML tags, URL lists
                      \item Repeated text blocks (e.g.\ duplicated paragraphs or sentences due to crawling errors)
                      \item Any peripheral, fragmented text clearly unrelated to the article's core subject.
                  \end{itemize}
            \item \textit{Judging Criteria:} If the absence of a certain text segment does not affect the article's logical coherence, the completeness of its core information, or the reader's understanding of the main topic, it can be considered useless information.
        \end{itemize}

  \item \textbf{Core Content \& Structure Retention (Of Paramount Importance):}
        \begin{itemize}[nosep, leftmargin=10pt]
            \item \textbf{Ensure that all key information, arguments, facts, data, explanations, examples, and logical deductions from the original text are completely intact.} No form of summarization, generalization, paraphrasing, or information loss is permitted.
            \item \textbf{Maintain the original narrative structure:} Retain original paragraph divisions, heading levels (e.g.\ H1, H2, etc.), and list formats (if any, e.g.\ ordered/unordered lists) as much as possible, unless these structures themselves constitute crawler traces.
            \item \textbf{Retain the original ``human touch'':} Preserve the author's original tone, style, and textual fluidity, avoiding any machine-generated or overly rigid sound.
        \end{itemize}

  \item \textbf{Text Coherence and Fluency:}
        \begin{itemize}[nosep, leftmargin=10pt]
            \item The refined text must be a self-contained, complete, and logically coherent whole.
            \item After removing useless information, transitions in the text should be natural and smooth, with no abruptness or disconnections.
            \item Ensure the text remains grammatically correct and clearly expressed after redundancy removal.
        \end{itemize}

  \item \textbf{No Repetition, High Precision:}
        \begin{itemize}[nosep, leftmargin=10pt]
            \item The final text should not contain any duplicated sentences or paragraphs.
            \item The quality of the generated text must meet the extremely high standards of an SFT answer, meaning accuracy, completeness, and usability approaching 100\%.
        \end{itemize}

  \item \textbf{Robustness:}
        \begin{itemize}[nosep, leftmargin=10pt]
            \item If the input raw text is already very clean and does not contain any of the elements listed above for removal, please return its content directly encapsulated in the JSON format, without any modifications.
        \end{itemize}
\end{enumerate}

\texttt{[Input Text]}: \textcolor{red}{\textbf{\{text\}}}\\\\
\texttt{[Output Format]} \\
\{ \\
\hspace*{2mm}``text'': ``refined text''\\
\}
\end{promptbox}
\caption{Prompt Template for Text Refinement.}
\label{fig:hss_text_refinement_prompt}
\end{figure*}

\begin{figure*}[t]
\centering
\begin{promptbox}{Prompt Template for {Refinement Evaluation}}
You are a text quality assessment expert specializing in evaluating answer texts. Your task is to carefully compare the original document and the refined document, evaluating whether the refinement meets the requirements for a high-quality answer.\\

\texttt{[Original Document (Seed Document)]}: \textcolor{red}{\textbf{\{text\}}}\\

\texttt{[Refined Document]}: \textcolor{red}{\textbf{\{refined text\}}}\\

\texttt{[Evaluation Task]}
Please evaluate the performance of the ``Refined Document'' relative to the ``Original Document'' based on the following criteria:\\
\begin{enumerate}[topsep=0pt, itemsep=4pt, parsep=0pt, leftmargin=14pt]
  \item \textbf{Redundant Information Removal}:  
  Has the refined text thoroughly removed all crawler traces and redundant information (e.g., web navigation, advertisements, copyright notices, excessive blank lines, duplicate content, fragmented text not central to the core topic, etc.)?

  \item \textbf{Core Content and Structure Integrity}:  
  Has the refined text 100\% preserved all core content, key information, data, arguments, narrative structure, and the author's tone and ``human touch'' from the original text, without any form of summarization, generalization, rephrasing, or information loss?

  \item \textbf{Text Coherence and Quality}:  
  Is the refined text natural and fluent, without abruptness or breaks, grammatically correct, and does it meet the extremely high standards of accuracy, completeness, and usability required for an SFT answer?

  \item \textbf{Robustness (if original text is clean)}:  
  If the ``Original Document'' itself was already very clean, did the refined text remain completely unchanged, without unnecessary modifications?
\end{enumerate}

\texttt{[Scoring (1-5)]}  
\begin{itemize}[topsep=2pt, itemsep=4pt, parsep=0pt, leftmargin=12pt]
  \item 1: Very poor – completely fails to meet requirements, severe issues, unusable as an SFT answer.
  \item 2: Poor – multiple issues, unsuitable as an SFT answer.
  \item 3: Fair – meets some requirements, but still has significant room for improvement.
  \item 4: Good – largely meets requirements, with only minor, negligible issues.
  \item 5: Excellent – perfectly meets all requirements, can directly serve as a high-quality SFT answer.
\end{itemize}

\texttt{[Output Format]} \\
\{ \\
\hspace*{2mm}``Reason'': ``<detailed explanation>'', \\
\hspace*{2mm}``Score'': <1-5> \\
\}

\end{promptbox}
\caption{Prompt Template for Refinement Evaluation.}
\label{fig:hss_refinement_evaluation_prompt}
\end{figure*}

\begin{figure*}[t]
\centering
\begin{promptbox}{Prompt Template for {Instruction Inversion}}
You are an expert in instruction inversion. Your task is to generate a ``reverse synthesis instruction'' for a given text.\\

\texttt{\# GUIDELINES FOR INSTRUCTION GENERATION}\\

\texttt{\textbf{Part 1: The Core Instruction}}\\
This section outlines ``what'' to the task — the specific, actionable requirements for generating the instruction.

\begin{enumerate}[topsep=2pt, itemsep=4pt, parsep=0pt, leftmargin=14pt]
  \item \textbf{Objective}: The generated instruction should explicitly guide the writer to create a piece of writing that is highly similar to the given text.
  \item \textbf{Domain}: The instruction must clearly specify the text's \textbf{Domain} as \textcolor{red}{\textbf{\{domain\}}} and \textbf{Genre} as \textcolor{red}{\textbf{\{genre\}}}.
  \item \textbf{Core Content \& Key Points}:
        \begin{itemize}[nosep, leftmargin=10pt]
          \item Summarize the central topic and the main thesis/argument of the text.
          \item List the essential supporting points, key data, primary examples, or narrative events that \textbf{must} be included to preserve the original's substance.
          \item Ensure all essential elements-such as key figures, significant events, central concepts, and defining terms—must be retained with exactness.
        \end{itemize}
  \item \textbf{Structure \& Narrative Voice}:
        \begin{itemize}[nosep, leftmargin=10pt]
          \item Describe the text's organizational structure (e.g., problem-solution, chronological, thematic).
          \item Specify the \textbf{Narrative Voice} (e.g., first-person reflective, third-person objective narrator).
        \end{itemize}
  \item \textbf{Length Constraint}: Strictly enforce the word count. State clearly: ``The final text must be approximately \textcolor{red}{\textbf{\{token\_len\}}} tokens (around \textcolor{red}{\textbf{\{char\_len\}}} characters), matching the length of the original document.''
  \item \textbf{Concise Without Repetition}: The instruction sentence itself must be concise and avoid unnecessary repetition.
  \item \textbf{No Source Reference}: The instruction must not mention, imply, or hint to the ``source document'', ``original text'', etc.
\end{enumerate}

\texttt{\\\textbf{Part 2: The Persona}}\\
This section defines the ``how''—the persona the writer should adopt. It's a role-playing guide to capture the original author's spirit. Frame this entire section in the second person (``You are...'').

\begin{enumerate}[topsep=2pt, itemsep=4pt, parsep=0pt, leftmargin=14pt]
  \item \textbf{Stance}: Based on the text's content, describe the author's likely underlying stance or values when discussing the topic.
  \item \textbf{Mindset \& Tone}: Based on the text's overall emotional color and undertones, describe the author's probable underlying mindset and emotional tone.
\end{enumerate}

\texttt{\# INSTRUCTION GENERATION TASK}\\
Please generate the ``reverse synthesis instruction'' for the following document: \textcolor{red}{\textbf{\{text\}}}\\

\texttt{\# FINAL OUTPUT FORMAT}\\
Provide the output in a single JSON object with two keys: \texttt{``Instruction''} and \texttt{``Persona''}.

\{\\
\hspace*{2mm}"Instruction": "(The core task instruction generated according to Part 1 guidelines)",\\
\hspace*{2mm}"Persona": "(The persona generated according to Part 2 guidelines, written in the second person.)"\\
\}

\end{promptbox}
\caption{Prompt Template for Instruction Inversion.}
\label{fig:hss_instruction_inversion_prompt}
\end{figure*}

\begin{figure*}[t]
\centering
\begin{promptbox}{Prompt Template for {Q\&A Consistency Check}}

You are an expert in the quality evaluation of Q\&A pairs. Your task is to evaluate the quality of an inverted instruction. Evaluate whether the inverted instruction can guide a model to generate inverted text similar to the original text. According to the Instruction Effectiveness in the Evaluation Criteria, and using Textual Consistency as a supplementary reference, determine whether the inverted instruction is ``qualified'' or ``unqualified''. Your final output is a binary conclusion: \textbf{1 means qualified}, \textbf{0 means unqualified}.

\texttt{[Input]}\\
INVERTED INSTRUCTION: The instruction to be evaluated.\\
ORIGINAL TEXT: The human-written reference (``correct answer'').\\
INVERTED TEXT: The actual output generated by the model based on the inverted instruction.\\

\verb|<INVERTED INSTRUCTION>| \textcolor{red}{\textbf{\{inverted\_instructions\}}} \verb|</INVERTED INSTRUCTION>|\\
\verb|<ORIGINAL TEXT>| \textcolor{red}{\textbf{\{text\}}} \verb|</ORIGINAL TEXT>|\\
\verb|<INVERTED TEXT>| \textcolor{red}{\textbf{\{generated\_text\}}} \verb|</INVERTED TEXT>|

\texttt{[Evaluation Criteria]}\\
Please judge whether the inverted instruction is qualified strictly according to the following standards.

\begin{itemize}[topsep=2pt, itemsep=4pt, parsep=0pt, leftmargin=12pt]
  \item \textbf{1 (Qualified)}
        \begin{itemize}[leftmargin=*, labelsep=0.5em, itemsep=2pt]
          \item \textbf{Textual Consistency}: The core content of the inverted text is strictly consistent with the original text.
          \item \textbf{Instruction Effectiveness}: The inverted instruction should include the core content from the original text. The inverted instruction must NOT contain words or phrases such as ``original'' or ``the above content'' that directly refer to the original text.
        \end{itemize}

  \item \textbf{0 (Unqualified)}
        \begin{itemize}[leftmargin=*, labelsep=0.5em, itemsep=2pt]
          \item \textbf{Textual Inconsistency}: The core content of the inverted text is inconsistent with the original text. The key characters, key events, or key messages are inconsistent.
          \item \textbf{Instruction Ineffectiveness}: The inverted instruction lacks the core content and key characters from the original text, or contains incorrect or fabricated information. The inverted instruction contains words or phrases such as ``original'' or ``the above content'' that directly refer to the original text.
        \end{itemize}
\end{itemize}

\texttt{[Output Format]}\\
Return a JSON object with two keys:

\{\\
\hspace*{2mm}"reason": "An explanation of why you chose the score",\\
\hspace*{2mm}"score": <1 \textbar{} 0>\\
\}

\end{promptbox}
\caption{Prompt Template for Q\&A Consistency Check.}
\label{fig:hss_qa_consistency_check_prompt}
\end{figure*}

\begin{figure*}[t]
\centering
\begin{promptbox}{Prompt Template for {Controlled Quality Degradation}}
You are an expert in text simplification.  
Based on previous expert reviews of the text, your task is to perform a \textbf{controlled simplification} of the original text.  
The primary goal is to reduce its performance level across four dimensions: \textbf{``Literary Diversity'', ``Theme Exploration Depth'', ``Emotional Expression'', and ``Humanities Creativity''} (based on provided current and target scores), while strictly adhering to all the requirements listed below.

\textbf{\# Operational Guides for Quality Degradation}
\begin{enumerate}[topsep=2pt, itemsep=4pt, parsep=0pt, leftmargin=14pt]
  \item \textbf{Literary Diversity} \\
  Current Situation: \textcolor{red}{\{Literary Diversity\}} \\
  Target Situation: \textcolor{red}{\{Literary Diversity Degrade\}} \\
  \textit{Operating Guide:}
  \begin{itemize}[nosep, leftmargin=10pt]
    \item Streamline rhetorical devices, use more direct expression.
    \item Simplify sentence structures, avoid overly ornate forms.
    \item Use a single narrative perspective, avoid perspective shifts.
  \end{itemize}

  \item \textbf{Theme Exploration Depth} \\
  Current Situation: \textcolor{red}{\{Theme Exploration Depth\}} \\
  Target Situation: \textcolor{red}{\{Theme Exploration Depth Degrade\}} \\
  \textit{Operating Guide:}
  \begin{itemize}[nosep, leftmargin=10pt]
    \item Retain core points but reduce in-depth argumentation.
    \item Summarize examples, keep only necessary factual details.
  \end{itemize}

  \item \textbf{Emotional Expression} \\
  Current Situation: \textcolor{red}{\{Emotional Expression\}} \\
  Target Situation: \textcolor{red}{\{Emotional Expression Degrade\}} \\
  \textit{Operating Guide:}
  \begin{itemize}[nosep, leftmargin=10pt]
    \item Moderately reduce strong tones or subjective language.
    \item Move toward a restrained, neutral tone.
  \end{itemize}

  \item \textbf{Humanities Creativity} \\
  Current Situation: \textcolor{red}{\{Humanities Creativity\}} \\
  Target Situation: \textcolor{red}{\{Humanities Creativity Degrade\}} \\
  \textit{Operating Guide:}
  \begin{itemize}[nosep, leftmargin=10pt]
    \item Use conventional expressions and common argument structures.
    \item Reduce unique or experimental narrative structures.
  \end{itemize}
\end{enumerate}

\textbf{\# Overall Requirements}
\begin{enumerate}[topsep=2pt, itemsep=4pt, parsep=0pt, leftmargin=14pt]
  \item Ensure grammar, punctuation, and spelling are correct.
  \item Maintain logical coherence and smooth transitions.
  \item Ensure information accuracy and relevance to the core theme.
  \item \textbf{Strictly avoid repeating the same ideas or key facts.}
  \item Use relatively diverse vocabulary; avoid repetitive wording.
  \item Preserve necessary background and domain knowledge.
  \item Retain the tone and style appropriate to \textcolor{red}{\{Genre\}}.
  \item Ensure the text remains highly relevant to \textcolor{red}{\{Domain\}}.
  \item Keep core arguments, key information, and critical elements.
  \item Simplify by refining and integrating content, not by deleting indiscriminately.
\end{enumerate}

\textbf{\# Length Specification}
This is a \textcolor{red}{\{length\}} token document.
The simplified text should be approximately or more than \textcolor{red}{\{target\_length\}} words.\\
\\

\textbf{Input Format:} \textcolor{red}{\textbf{\{text\}}}\\
\textbf{Output Format:}
\begin{verbatim}
{"text": "simplified text"}
\end{verbatim}
\end{promptbox}
\caption{Prompt Template for Controlled Quality Degradation.}
\label{fig:hss_controlled_quality_degradation_prompt}
\end{figure*}

\end{document}